\documentclass[acmtog]{acmart}
\usepackage[ruled,vlined]{algorithm2e}
\usepackage{hyperref}
\usepackage{amsmath}
\usepackage{mathrsfs}
\usepackage{bbding}
\RestyleAlgo{ruled}
\RestyleAlgo{vlined}
\acmJournal{TOG}

\usepackage{soul}
\usepackage{booktabs}
\usepackage{pifont}

\SetKwComment{Comment}{// }{}

\newcommand{\ie}{i.\,e.\ }	
\newcommand{\eg}{e.\,g.\ }	

\newcommand{\revised}[1]{\textcolor{black}{#1}}

\def\ie{\emph{i.e.,~}}
\def\eg{\emph{e.g.,~}}

\newcommand{\figref}[1]{Figure~\ref{#1}}%
\newcommand{\tabref}[1]{Table~\ref{#1}}%
\newcommand{\secref}[1]{Section~\ref{#1}}
\renewcommand{\eqref}[1]{Equation.~(\ref{#1})}

\definecolor{myyellow}{rgb}{1, 1, 0.7}
\definecolor{myorange}{rgb}{1, 0.85, 0.7}
\definecolor{myred}{rgb}{1, 0.7, 0.7}
\newcommand{\reducedstrut}{\vrule width 0pt height 1.05\ht\strutbox depth 1.0\dp\strutbox\relax}
\newcommand{\sota}[1]{%
  \begingroup
  \setlength{\fboxsep}{0pt}%
  \colorbox{myred}{\reducedstrut#1\/}%
  \endgroup
}
\newcommand{\subsota}[1]{%
  \begingroup
  \setlength{\fboxsep}{0pt}%
  \colorbox{myorange}{\reducedstrut#1\/}%
  \endgroup
}
\newcommand{\third}[1]{%
  \begingroup
  \setlength{\fboxsep}{0pt}%
  \colorbox{myyellow}{\reducedstrut#1\/}%
  \endgroup
}

\usepackage{graphicx}
\usepackage{subcaption}
\usepackage{multirow}
\usepackage{makecell, multicol, diagbox}
\usepackage{xcolor}
\usepackage{pdfpages}

\definecolor{zoomin}{RGB}{95, 151, 210}
\definecolor{floater}{RGB}{50,184,151}

\makeatletter
\newcommand\paragraphNew{\@startsection{paragraph}{4}{\parindent}%
  {-.5\baselineskip \@plus -2\p@ \@minus -.2\p@}%
  {-3.5\p@}%
  {\ACM@NRadjust{\@parfont}}}
	\makeatother
	
\AtBeginDocument{%
  \providecommand\BibTeX{{%
    \normalfont B\kern-0.5em{\scshape i\kern-0.25em b}\kern-0.8em\TeX}}}

\acmJournal{TOG}

\setcopyright{acmlicensed}
\acmJournal{TOG}
\acmYear{2025} \acmVolume{1} \acmNumber{1} \acmArticle{1} \acmMonth{1}\acmDOI{10.1145/3759248}

\begin{document}


\title{GS-ROR$^2$: Bidirectional-guided 3D\textbf{GS} and SDF for \textbf{R}eflective \textbf{O}bject \textbf{R}elighting and \textbf{R}econstruction}


\author{Zuo-Liang Zhu}
\affiliation{
 \institution{VCIP, College of Computer Science, Nankai University}
 \country{China}
}
\email{nkuzhuzl@gmail.com}

\author{Beibei Wang$^\dagger$}
\thanks{$^\dagger$Corresponding author.}
\affiliation{
   \institution{School of Intelligence Science and Technology,  Nanjing University}
   \country{China}
}
\email{beibei.wang@nju.edu.cn}

\author{Jian Yang$^\dagger$}
\affiliation{
  \institution{VCIP, College of Computer Science, Nankai University}
  \country{China}
}
\email{csjyang@nankai.edu.cn}


\begin{abstract}

3D Gaussian Splatting (3DGS) has shown a powerful capability for novel view synthesis due to its detailed expressive ability and highly efficient rendering speed. Unfortunately, creating relightable 3D assets and reconstructing faithful geometry with 3DGS is still problematic, particularly for reflective objects, as its discontinuous representation raises difficulties in constraining geometries.
In contrary, volumetric signed distance field (SDF) methods provide robust geometry reconstruction, while the expensive ray marching hinders its real-time application and slows the training. Besides, these methods struggle to capture sharp geometric details. To this end, we propose to guide 3DGS and SDF bidirectionally in a complementary manner, including an SDF-aided Gaussian splatting for efficient optimization of the relighting model and a GS-guided SDF enhancement for high-quality geometry reconstruction. At the core of our SDF-aided Gaussian splatting is the \emph{mutual supervision} of the depth and normal between blended Gaussians and SDF, which avoids the expensive volume rendering of SDF. Thanks to this mutual supervision, the learned blended Gaussians are well-constrained with a minimal time cost. 
As the Gaussians are rendered in a deferred shading mode, the alpha-blended Gaussians are smooth, while individual Gaussians may still be outliers, yielding floater artifacts. Therefore, we introduce an SDF-aware pruning strategy to remove Gaussian outliers located distant from the surface defined by SDF, avoiding the floater issue. This way, our GS framework provides reasonable normal and achieves realistic relighting, while the mesh of truncated SDF (TSDF) fusion from depth is still problematic. Therefore, we design a GS-guided SDF refinement, which utilizes the blended normal from Gaussians to finetune SDF. 
Equipped with the efficient enhancement, our method can further provide high-quality meshes for reflective objects at the cost of 17\% extra training time. Consequently, our method outperforms the existing Gaussian-based inverse rendering methods in terms of relighting and mesh quality. 
Our method also exhibits competitive relighting/mesh quality compared to NeRF-based methods with at most 25\%/33\% of training time and allows rendering at 200+ frames per second on an RTX4090. Our code is available at \url{https://github.com/NK-CS-ZZL/GS-ROR}.

\end{abstract}

\ccsdesc[500]{Computing methodologies~Rendering}

\keywords{neural rendering, Gaussian splatting, relighting}

\begin{teaserfigure}
\centering
\includegraphics[width=\textwidth]{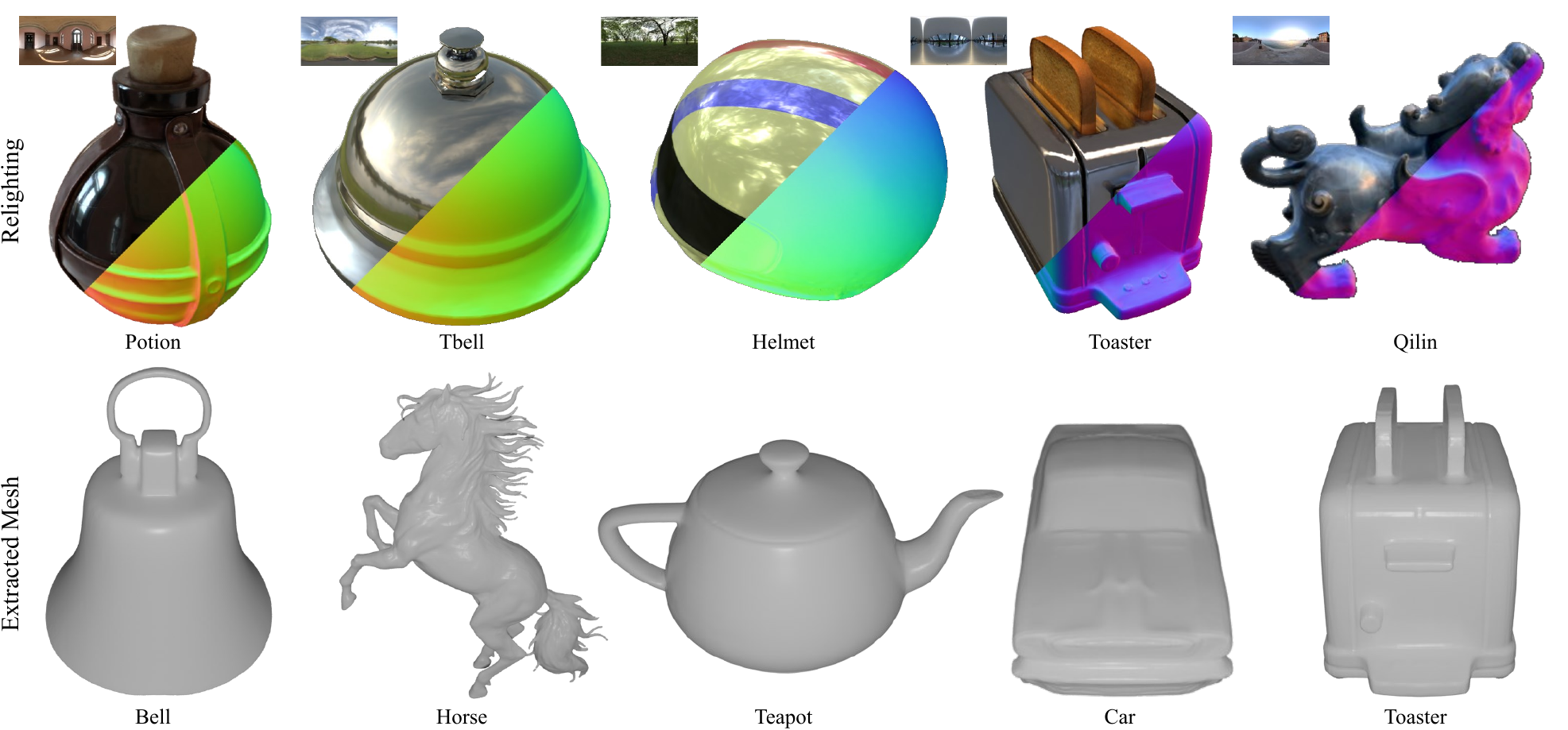}
\Description[The teaser figure]{Representative relighting results and mesh visualizations of reflective objects from our method, including realistic rendering and reasonable geometry.}
\caption{We present bidirectional-guided 3D\textbf{GS} and SDF for \textbf{R}eflective \textbf{O}bject \textbf{R}elighting and \textbf{R}econstruction (GS-ROR$^2$) from multi-view images. 
We show relighting results with reflective highlights (left) and their normal (right) in the $1^{st}$ row, including \textsc{Potion}, \textsc{Tbell} from NeRO~\cite{liu_2023_nero}, \textsc{Helmet}, \textsc{Toaster} from Ref-NeRF~\cite{verbin_2022_refnerf}, and \textsc{Qilin} from NeILF++~\cite{zhang_2023_neilfpp}, where \textsc{Qilin} is a real scene. Besides, we show extracted meshes in the $2^{nd}$ row, including \textsc{Bell}, \textsc{Horse}, \textsc{Teapot} from NeRO and \textsc{Car}, \textsc{Toaster} from Ref-NeRF. Our method demonstrates a robust geometry reconstruction for reflective surfaces and faithful material decomposition, leading to photo-realistic reflective object relighting and high-fidelity geometry reconstruction. 
}
\label{fig:teaser}
\end{teaserfigure}


\maketitle

\section{Introduction}

Creating relightable 3D assets from multi-view images has been a long-standing and challenging task in computer graphics and vision, as the decomposition of lighting, materials, and geometries is highly ill-posed. Particularly, the decomposition becomes more difficult for reflective objects, as their appearances are highly view-dependent, and a minor error on the surface leads to a significant difference. Existing approaches~\cite{liu_2023_nero,li_2024_tensosdf} have shown impressive relighting quality for reflective objects by leveraging the neural radiance field (NeRF) and the signed distance field (SDF). Unfortunately, these methods require a long training and rendering time. They also struggle to capture sharp geometry details. In this paper, we aim at reflective object relighting and reconstruction given multi-view images, achieving high-quality relighting and detailed reconstruction with short training/rendering time costs. In \figref{fig:teaser}, we present some representative results from our method, including the relighting results and mesh visualizations.

Most recently, Kerbl et al.~\shortcite{kerbl_2023_3dgs} proposed 3D Gaussian Splatting (3DGS), boosting the rendering speed significantly and achieving more detailed appearance modeling. A concurrent work~\cite{ye_2024_3dgsdr} has introduced deferred Gaussian splatting for reflective objects, improving the novel view synthesis (NVS) quality. Despite the impressive NVS quality, relighting and geometry reconstruction with 3DGS become problematic due to the Gaussians' discontinuity and high flexibility, especially for reflective objects. Extensive efforts have been made to improve the relighting quality by regularizing the geometry in terms of the normal and depth~\cite{gao_2023_relightablegs, jiang_2024_gaussianshader, liang_2024_gsir}. However, they still suffer from erroneous surfaces and floater artifacts for reflective objects.

Some prior works (\eg NeuSG~\cite{chen_2023_neusg}, GSDF~\cite{yu_2024_gsdf}) couple SDF with GS for surface reconstruction. While their work has shown impressive reconstructed quality for diffuse objects, \revised{they require extensive training time}, due to the rendering of SDF and the existence of two representations. Our method focuses on both relighting and reconstruction, thus raising more difficulties. \revised{Specifically, evaluating the rendering equation increases the training time.} These issues become more severe when modeling reflective objects, as the specular highlight is sensitive to the reflective direction, which needs more Gaussians to provide detailed normal. Therefore, introducing SDF into GS for our task is not straightforward, which raises the demand for a more efficient framework.

In this paper, we propose a bidirectional guidance of GS and SDF in a complementary manner, including an SDF-aided Gaussian splatting for efficient optimization of the relighting model and a GS-guided SDF enhancement for high-quality geometry reconstruction. Specifically, the SDF-aided Gaussian splatting utilizes the SDF to constrain the Gaussians with minimal computational increase. For this, we propose to supervise the deferred Gaussians and a low-resolution SDF mutually in our framework, so-called \emph{mutual supervision}, which allows constraining deferred Gaussians with SDFs while avoiding SDF rendering. This way, the learned deferred Gaussians are better constrained. However, we still notice that Gaussians have floater artifacts occasionally, as the deferred Gaussians (with alpha blending) rather than individual Gaussians are constrained by SDF. Therefore, we introduce an SDF-aware pruning strategy to remove Gaussian outliers to avoid this floater issue. Therefore, our method provides reasonable shading normal and enables realistic relighting for reflective objects in the GS framework. Due to the inconsistency between the normal and depth from blended Gaussians, the meshes of TSDF fusion from depth are still problematic. Meanwhile, SDF is underfitted and only learns the coarse structure of geometry. To solve this, we further design a GS-guided SDF enhancement, which utilizes the learned normal from fixed Gaussians to finetune the upsampled SDF. This way, we avoid optimizing a full-resolution SDF and GS jointly, leading to an efficient refinement. SDF benefits from the detailed expressive ability of GS and learns more fine-grained geometry details. Equipped with this design, our method achieves high-quality geometry reconstruction for reflective objects from SDF with only 17\% extra overhead of training time. Consequently, our method outperforms the existing Gaussian-based inverse rendering methods in terms of relighting and mesh quality. Meanwhile, it exhibits competitive relighting/mesh quality compared to NeRF-based methods with at most 25\%/33\% of training time. It also allows real-time rendering with 200+ frames per second on an RTX4090. To summarize, our main contributions include 
\begin{itemize}
\item a GS-ROR$^2$ framework for reflective objects, achieving high-quality relighting and geometry reconstruction while maintaining short training time and allowing real-time rendering,  
\item an SDF-aided GS, including a mutual supervision between deferred Gaussians and the SDF with an SDF-aware Gaussian pruning strategy, producing robust geometry and avoiding local minima with minimal computational increase, and
\item a GS-guided SDF enhancement via blended normal from Gaussians, enabling high-quality geometry reconstruction.
\end{itemize}

\section{Related Work}
\subsection{Neural representations for multi-view stereo}
Since the presence of NeRF~\cite{mildenhall_2020_nerf}, numerous neural implicit representations~\cite{fridovich_2022_plenoxels, barron_2022_mipnerf360, muller_2022_ingp, chen_2022_tensorf} have been proposed and gained remarkable progress in the field of multi-view stereo. These NeRF-based models adopt Multi-Layer Perceptrons (MLPs)~\cite{barron_2021_mipnerf, barron_2022_mipnerf360, verbin_2022_refnerf} or grid-like representations~\cite{muller_2022_ingp, chen_2022_tensorf, fridovich_2022_plenoxels} to represent geometry and view-dependent appearance, optimized via volume rendering. Subsequently, NeuS~\cite{wang_2021_neus} links the SDF with the density in NeRF and enables the optimization of surfaces with volume rendering. With extra regularization, these SDF-based models~\cite{li_2023_neuralangelo, rosu_2023_permutosdf, wang_2023_neuralsingular} achieve detailed surface reconstruction from RGB images. Owing to carefully designed architectures and specific optimization, some methods~\cite{muller_2022_ingp, fridovich_2022_plenoxels} enable fast training and real-time radiance field rendering.   
%

3DGS~\shortcite{kerbl_2023_3dgs} sheds light on the multi-view stereo in the rasterization framework, providing impressive real-time NVS results. 3DGS utilizes the discrete explicit 3D Gaussian primitive and projects these primitives onto the image plane to obtain the pixel color via alpha blending, which avoids the consuming sampling operation. However, 3DGS imposes no geometry constraint while optimizing, leading to noisy depth and normal, which hinders the usage of 3DGS in many downstream tasks. To resolve this issue, existing methods construct different constraints on Gaussians by introducing 2D Gaussian~\cite{huang_2024_2dgs} or Gaussian Surfel~\cite{dai_2024_gaussiansurfels}. SuGaR~\cite{guedon_2024_sugar} extracts meshes from the Gaussians and binds the flattened Gaussians to the surface to optimize further. These new primitives and regularization could improve the geometry quality, potentially benefiting the relighting task.


Several previous works introduce SDF into 3DGS. NeuSG introduces SDF into 3DGS to improve the details of surface reconstruction by optimizing NeuS and 3DGS jointly. Despite the high quality of the reconstructed surface, the training time is tens of times longer than 3DGS. GSDF utilizes the depth from Gaussian to guide sampling for SDF, but they still need to render color in the SDF branch, leading to higher efficiency. While our method also incorporates the SDF and Gaussians, the key difference is that it drops SDF rendering, improving the geometry quality with a low time cost.



\subsection{Inverse rendering}
Inverse rendering aims to decompose geometry, material, and light from multi-view RGB images. Most existing methods adopt the geometry formulation from NeRF in either density or SDF manner, namely NeRF-based method. In terms of material modeling, most methods utilize the Disney Principled BRDF model~\cite{burley_2012_physically}, while NeRFactor~\cite{zhang_2021_nerfactor} pretrains a BRDF MLP with a measured BRDF dataset. 
Considering the light modeling, early works~\cite{zhang_2021_physg, boss_2021_neuralpil, bi_2020_neural, boss_2021_nerd, zhang_iron_2022} consider direct light only, and recent works~\cite{srinivasa_2021_nerv, zhang_2022_mii, jin_2023_tensoir, zhang_2023_neilfpp, yang_2023_sireir, sun_2024_neuralpbir} model indirect lighting and visibility, leading to a higher-quality decomposition. NeRF Emitter~\cite{liu_2024_nerfemitter} incorporates NeRF as a non-distant emitter and uses importance sampling of NeRF to reduce rendering variance. Neural-PBIR~\cite{sun_2024_neuralpbir} introduces the differentiable path tracing to model complex inter-reflection.
However, these NeRF-based methods suffer from slow rendering speed. 

More recently, Gaussian-based inverse rendering methods have begun to emerge. 
These methods regularize Gaussians to obtain better geometry in various ways. GShader~\cite{jiang_2024_gaussianshader} and GIR~\cite{shi_2023_gir} link the shortest axis of Gaussian with the normal. The GShader, R3DG~\cite{gao_2023_relightablegs}, and GS-IR~\cite{liang_2024_gsir} supervise the Gaussian normal with the depth normal. 
To model the indirect illumination, R3DG, GIR, and GS-IR compute the visibility via one-step ray tracing and represent the indirect illumination with SH coefficients. These early Gaussian-based relighting methods cannot relight reflective objects, while our method enables such scenarios.

\subsection{Reflective objects NVS and relighting}
As a special and challenging case, NVS and relighting of reflective objects have drawn much attention. 
Ref-NeRF~\cite{verbin_2022_refnerf} replaces the outgoing radiance with the incoming radiance and models the shading explicitly with
an integrated direction encoding (IDE) to improve the quality of reflective materials. MS-NeRF~\cite{yin_2023_msnerf} designs a multi-space neural representation to model the reflection of mirror-like planes. Spec-NeRF~\cite{ma_2024_specnerf} introduces a learnable Gaussian directional encoding to better model view-dependent effects. Spec-Gaussian~\cite{yang_2024_specgaussian} uses anisotropic spherical Gaussian to model view-dependent highlights. 3DGS-DR~\cite{ye_2024_3dgsdr}, as a concurrent work, introduces deferred shading into 3DGS. We also utilize a deferred framework, but for relighting, and together with other key components.

Another group of methods enables relighting for reflective materials. NeRO~\cite{liu_2023_nero} proposes a novel light representation consisting of two MLPs for direct and indirect light, respectively. With explicit incorporation of the rendering equation, NeRO reconstructs reflective objects of high quality. Wang et al.~\shortcite{wang_2024_inverse} represent the indirect light with a 5-dimensional function, so-called the 5D neural plenoptic function. Thanks to material-aware integral positional encoding, their method reduces the light sample per pixel during training and provides more accurate material decomposition. TensoSDF~\cite{li_2024_tensosdf} designs roughness-aware incorporation between the radiance and reflectance field for robust geometry reconstruction and proposes a novel tensorial representation with SDF, enabling fast SDF query. Although their representation is faster than the original SDF, it still needs hours of training time and cannot achieve real-time rendering. Recently some Gaussian-based methods enable reflective object relighting with the help of pretrained models. DeferredGS~\cite{wu_2024_deferredgs} utilizes the idea of deferred shading to improve the quality of the reflective-object relighting in the Gaussian framework. However, their method needs to distill reasonable geometry from a pretrained SDF, taking more than 3 hours on an RTX 4090 and thus losing the efficiency in training. GlossyGS~\cite{lai_2024_glossygs} introduces the segmentation prior for a reasonable material estimation, while the segmentation model (i.e., DINOv2~\cite{oquab_2024_dinov2}) needs more than 3 days on 4 RTX A100 to finetune with custom data. Our method takes a further step in the reflective object relighting with 3D Gaussians and provides high-quality relighting results without any pretrained model, keeping highly efficient training and rendering. Furthermore, our method can provide high-quality mesh with efficient finetuning, while the existing Gaussian-based methods fail to provide high-quality meshes for reflective objects.

\begin{figure}[tb]
    \centering
    \includegraphics[width = \linewidth]{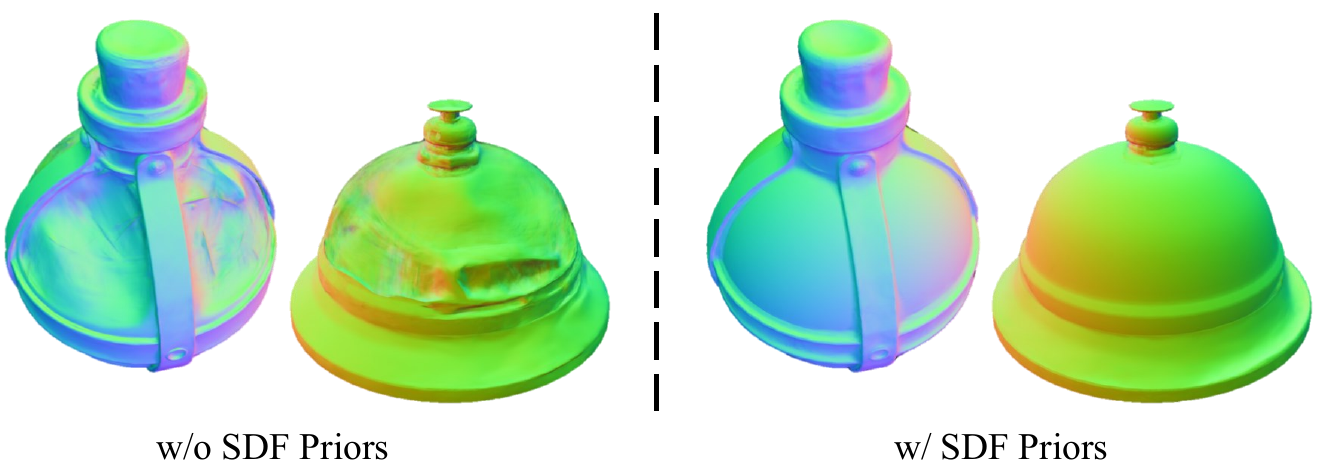}
    \caption{The geometry from Gaussian is under-constrained and thus erroneous, while it is much better after utilizing the priors from the SDF.}
    \Description[Comparison of normals with and without SDF priors]{The predicted normals with SDF priors reveal to be reasonable, while the ones without the priors are erroneous.}
    \label{fig:motivation}
\end{figure}

\begin{figure*}[htb]
    \centering
    \includegraphics*[clip, width = \linewidth]{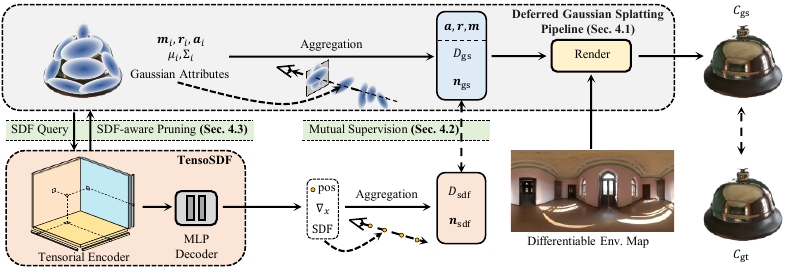}
    \caption{Overview of our SDF-aided Gaussian Splatting. The architecture of our proposed method consists of two geometry representations (\ie Gaussian primitive and TensoSDF). 
    In the deferred Gaussian pipeline,
    the shading parameters (\ie albedo $\textbf{a}$, roughness $\textbf{r}$ and metallicity $\textbf{m}$), normal and depth are projected to the image plane and alpha blended. The pixel color $C_{\rm gs}$ is calculated using the split-sum approximation and supervised by ground truth color $C_{\rm gt}$.
    In the TensoSDF, we sample rays originated from camera center $\textbf{o}$ and view direction $\textbf{v}$ and query the SDF value and gradient for each point $p$ along the ray $\textbf{o}+t\textbf{v}$. The normal $\textbf{n}_{\rm sdf}$ and depth $D_{\rm sdf}$ are obtained via volume rendering, which is supervised mutually with the normal $\textbf{n}_{\rm gs}$ the depth $D_{\rm gs}$ from Gaussians.
    Note no color network is used in the SDF part, and only the geometry attributes are volume rendered.}
    \Description[Overview of our SDF-aided Gaussian Splatting]{The framework includes a deferred Gaussian splatting to render images supervised by ground-truth images and a TensoSDF to constrain the geometry.}
    \label{fig:overview}
\end{figure*}

\section{Preliminaries and motivation}
In this section, we briefly review two ways for 3D scene representations and then discuss our motivation. 

\paragraph{3D Gaussian Splatting}
3DGS represents a scene with a set of 3D Gaussians whose distribution is defined as
\begin{equation}
    G(x) = e^{-\frac{1}{2}(x-\mu)^T\Sigma^{-1}(x-\mu)},
\end{equation}
where $x$ is a position in the scene, $\mu$ is the mean of the Gaussian, and $\Sigma$ denotes the covariance matrix of the 3D Gaussian, which is factorized into a scaling matrix $S$ and a rotation matrix $R$ as $\Sigma=RSS^TR^T$. To render an image, 3DGS projects the 3D Gaussians onto the image plane and employs alpha blending on the sorted 2D Gaussians as 
\begin{equation}
    C = \sum^{n}_{i=0}{c_i\alpha_i\prod_{j=1}^{i-1}(1-\alpha_j)},
    \label{eq:gsblend}
\end{equation}
where $c_i$ is the color of each Gaussian, and $\alpha_i$ is given by evaluating a projected 2D Gaussian with covariance $\Sigma^\prime$ multiplied with a learned per-point opacity.

\paragraph{TensoSDF}
An SDF expresses a scene by encoding the distance from the surface of an object, which can be expressed implicitly by an MLP~\cite{wang_2021_neus} or in a hybrid way by combining explicit grids with a tiny MLP~\cite{li_2024_tensosdf}, which allows more efficient training and rendering. Depth and normal are two significant attributes of geometry. In SDF, the point-wise depth $D_i$ is defined as the distance from the camera to the sample, and the point-wise normal $\textbf{n}_i$ is the gradient of SDF $\nabla_x$ w.r.t. position $x$~\cite{zhang_2021_physg}. Given $n$ points along the ray $p_i=\textbf{o}+t\textbf{v}$ where $\textbf{o}$ is the camera origin and $\textbf{v}$ is the view direction, we can aggregate these point-wise attributes to obtain the attributes (\ie depth, normal) of the hit point of the surface as:
\begin{equation}
    D_{\rm sdf} = \sum^{n}_{i=0}w_iD_i, \textbf{n}_{\rm sdf} = \sum^{n}_{i=0}w_i\textbf{n}_i
    \label{eq:sdf_color}
\end{equation}
where $w_i$ is the weight of the $i$-th point derived from the SDF value.

SDF and NeRF can be unified by NeuS and VolSDF~\cite{wang_2021_neus, yariv_2021_volsdf}, which maps an SDF value to a density distribution. The distribution in NeuS is defined as: 
\begin{equation}
    \phi_s(s)=\gamma e^{-\gamma s}/(1 + e^{-\gamma s})^2,
    \label{eq:dpf}
\end{equation}
where $s$ is the SDF value and $\gamma$ is the inverse of standard deviation. Note that $\gamma$ is a trainable parameter, and $1/\gamma$ approaches zero as the network training converges.

Among the SDF-based methods, TensoSDF utilizes a tensorial representation for SDF, consisting of a tensorial encoder and MLP decoder, which is formulated as follows:
\begin{equation}
    V_p = v^X_k \circ M^{YZ}_k \oplus v^Y_k \circ M^{XZ}_k \oplus v^Z_k \circ M^{XY}_k, s = \Theta(V_p, p),
\end{equation}
where $v^m_k$ and $M^{\widetilde{m}}_k$ represent the $k$-th vector and matrix factors of their corresponding spatial axes $m$, and $\widetilde{m}$ denotes the two axes orthogonal to $m$ \eg $\widetilde{X}=YZ$. $\circ$ and $\oplus$ represent the element-wise multiplication and concatenation operations. $V_p$ is the latent vector from the tensorial encoder and then is decoded with the position $p$ by a tiny MLP $\Theta$ to get the SDF value $s$. The TensoSDF backbone enables faster convergence speed and low-cost SDF queries. Besides, the capacity of TensoSDF can be modulated efficiently through the interpolation of the tensorial grid.



\paragraph{Motivation}
The above two representations can both express a 3D scene, where the Gaussian representation is explicit and discontinuous, and the SDF is implicit (or hybrid) and continuous. These two representations lead to different rendering methods: splatting or rasterization for Gaussians and sphere tracing for SDF. While the Gaussian representation significantly boosts the rendering speed to a real-time level, it has difficulties constraining geometries, leading to inferior geometries, as shown in \figref{fig:motivation}. A straightforward idea is to incorporate these two representations by leveraging the SDF as a prior. However, rendering SDF is time-consuming, which hinders the benefits of 3DGS. Therefore, the essential problem is an elegant incorporation between the Gaussians and the SDF.

\section{Method}


This section presents our bidirectional-guided GS and SDF for reflective object relighting and reconstruction, named GS-ROR$^2$. First, we present an SDF-aided GS for relighting, including a deferred GS pipeline (\secref{sec:defer}), an SDF prior of mutual supervision for geometry regularization (\secref{sec:sdf}), and an SDF-aware pruning to refine the Gaussian distribution (\secref{sec:prune}). Finally, we present a GS-guided SDF enhancement for high-quality geometry reconstruction (\secref{sec:ft}). 
In \figref{fig:overview}, we show a brief overview of our method.
%
\subsection{Deferred Gaussian splatting for relighting}
\label{sec:defer}

Starting from GS-IR~\cite{liang_2024_gsir}, we first construct a forward Gaussian splatting baseline for relighting, using the Disney Principled BDRF model~\cite{burley_2012_physically} and environment map to model the material and light. We drop the occlusion for efficiency and modify the normal formulation following Gaussian Shader~\cite{jiang_2024_gaussianshader} as the combination of the shortest axis direction $u$ and an offset $n_\delta$ as $\textbf{n} = u + n_\delta$. The radiance from view \textbf{v} for each Gaussian is computed by the split-sum approximation, following previous work~\cite{jiang_2024_gaussianshader, liang_2024_gsir}, and then aggregated by alpha blending. With multi-view images as inputs, the reflectance parameters at each Gaussian can be learned, besides the opacity and the environment map, enabling relighting under a novel environment map. 

\begin{figure}[tb]
    \centering
    \includegraphics[width = \linewidth]{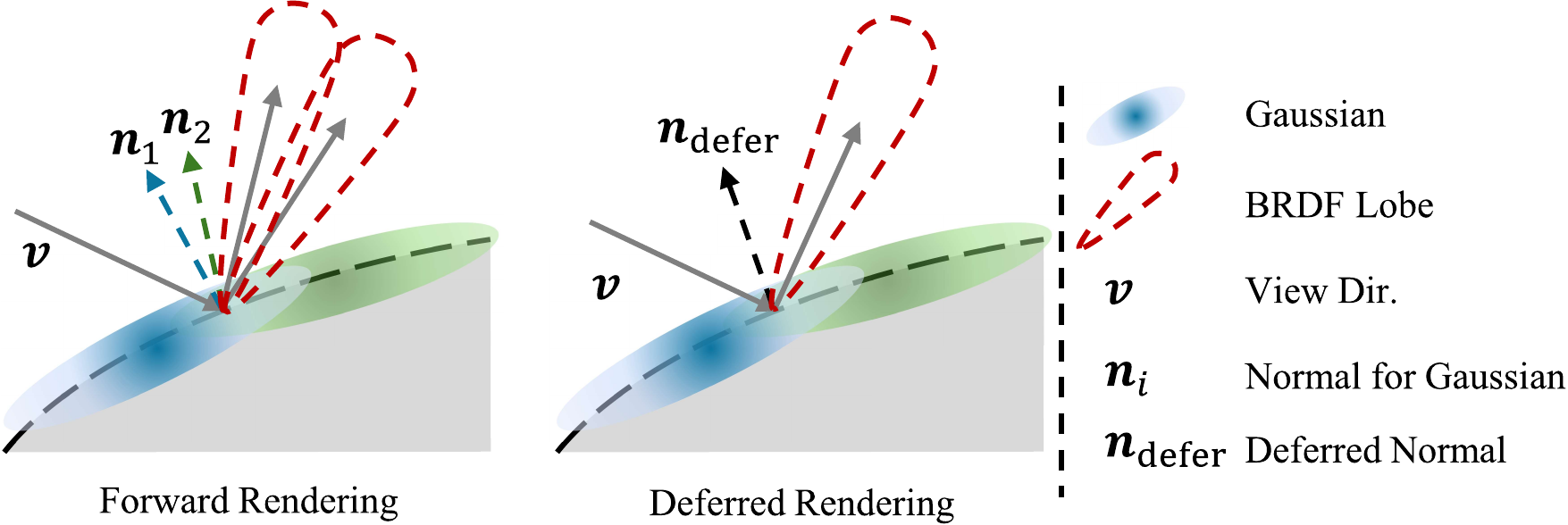}
    \caption{Two Gaussians with a minor normal difference are overlapped to model an opaque surface. In the forward rendering, the BRDF values are computed w.r.t. to their own normal and are then alpha-blended to form the final rendering, which is equivalent to a broader BRDF lobe, leading to a blurry rendering, eventually. In contrast, in the deferred shading, the BRDF is computed w.r.t. the deferred normal, maintaining the sharpness of reflective objects.}
    \Description[Deferred shading provides a sharp BRDF lobe.]{The forward shading used in vanilla 3D Gaussian splatting leads to a broader BRDF lobe and blurry rendering, while deferred shading Gaussian splatting maintains a sharp BRDF lobe.}
    \label{fig:defer}
\end{figure}

However, this simple Gaussian-based relighting baseline produces blurry relighting results for the reflective surface. The main reason behind this phenomenon is the alpha blending of Gaussians. As shown in \figref{fig:defer}, the Gaussians are overlapped to model an opaque surface with slightly different normals. While this difference among Gaussians causes a negligible error for rough objects, it amplifies the BRDF lobe for specular objects, leading to a blurry rendering. The key to addressing this issue is to blend the normals of Gaussians first and then perform shading, i.e., deferred shading.

Specifically, we perform alpha blending on all the attributes (\eg normal, albedo, roughness) onto the image plane:
\begin{equation}
    F = \sum^n_{i=0}{f_i\alpha_i\prod_{j=1}^{i-1}(1-\alpha_j)},
\end{equation}
where $\alpha_i, \alpha_j$ share the same definition as in \eqref{eq:gsblend}, and $f_i$ is a attribute of the $i$-th Gaussian. Then, we use split-sum approximation to compute the pixel-wise color. 

The Gaussians are learned with the following loss function:
\begin{equation}
    \mathcal{L}_{\rm gs} =\mathcal{L}_{\rm c} + \lambda_{\rm nd}\mathcal{L}_{\rm nd} + \lambda_{\rm sm}\mathcal{L}_{\rm sm} + \lambda_{\rm m}\mathcal{L}_{\rm m} + \lambda_{\rm \delta n}\mathcal{L}_{\rm \delta n},
    \label{eq:l_gs}
\end{equation}
where $\mathcal{L}_{\rm c}$ is the color supervision between the rendered image and ground truth as in 3DGS,  $\mathcal{L}_{\rm nd}$ is the supervision between the normal from Gaussian and from the estimated depth proposed in Gaussian Shader~\cite{jiang_2024_gaussianshader}, $\mathcal{L}_{\rm sm}$ is the smoothness loss for BRDF parameters, $\mathcal{L}_{\rm m}$ is the mask loss, and $\mathcal{L}_{\rm \delta n}$ is the delta normal regularization in Gaussian Shader. More details are shown in \secref{sec:opt}.

\subsection{An rendering-free mutual supervision}
\label{sec:sdf}

Thanks to the deferred Gaussian splatting, the blurry issue for reflective object relighting has been addressed. However, the geometry established by Gaussians exhibits erroneous surfaces and floaters, as it tends to overfit the appearance under training light conditions. Hence, more constraints on the geometry are needed. Inspired by NeRO and TensoSDF, we introduce SDF into our framework, as it has been shown to be a good prior for geometry regularization. The key is leveraging the capability of SDF while avoiding the expensive rendering. For this, we first choose a performance-friendly backbone for SDF -- TensoSDF.  Then, we propose geometry-only supervision between Gaussians and SDF, or so-called \emph{mutual supervision}, together with the color supervision between Gaussian renderings and the input images, without any SDF rendering. On the one hand, as Gaussians build the link between rendered images and input images, they can supervise the training of SDF. On the other hand, the SDF can also constrain the Gaussians.


\paragraph{Mutual supervision}
Even with a performance-friendly SDF backbone, TensoSDF rendering is still expensive, due to the complex color network required to faithfully represent the specular highlights. Therefore, we propose to remove the SDF rendering and connect the SDF and Gaussians with geometry attributes.

Specifically, we pose supervision on the depth and normal of the SDF (\eqref{eq:sdf_color}) from Gaussians (\eqref{eq:gsblend}):
\begin{equation}
\begin{split}
    \mathcal{L}_{\rm gs2sdf} = |\textbf{sg}[D_{\rm gs}] - D_{\rm sdf}| + (1 - <\textbf{sg}[\textbf{n}_{\rm gs}], \textbf{n}_{\rm sdf}>),
\end{split}
\label{eq:sdf2gs}
\end{equation}
where $D_{\rm gs}$ and $D_{\rm sdf}$ are the depth from Gaussians and SDF, respectively, $\textbf{n}_{\rm gs}$ and $\textbf{n}_{\rm sdf}$ are normals from Gaussians and SDF respectively, and $\textbf{sg}[\cdot]$ is the stop-gradient operation.

Besides the above loss, we also utilize the Eikonal loss $\mathcal{L}_{\rm eik}$, Hessian loss $\mathcal{L}_{\rm hes}$, mask loss $\mathcal{L}_{\rm m}$, and total variance loss $\mathcal{L}_{\rm tv}$ to regularize the training of SDF, following common practice~\cite{li_2024_tensosdf, gropp_2020_eik, zhang_2022_hes}. The final loss for SDF is 
\begin{equation}
    \mathcal{L}_{\rm sdf} = \lambda_{\rm gs2sdf}\mathcal{L}_{\rm gs2sdf} + \lambda_{\rm eik}\mathcal{L}_{\rm eik} + \lambda_{\rm hes}\mathcal{L}_{\rm hes} + \lambda_{\rm m}\mathcal{L}_{\rm m} + \lambda_{\rm tv}\mathcal{L}_{\rm tv},
    \label{eq:l_sdf}
\end{equation}
where the $\lambda_{[\cdot]}$ is the corresponding coefficient to adjust the strength of regularization. Among these losses, our proposed $\mathcal{L}_{\rm gs2sdf}$ enables the optimization of SDF via the geometry of Gaussians, and the $\mathcal{L}_{\rm eik}, \mathcal{L}_{\rm hes},\mathcal{L}_{\rm m},\mathcal{L}_{\rm tv}$ regularize the SDF to obtain smooth and reasonable geometry.

Then, we design another loss to update Gaussians with SDF priors:
\begin{equation}
    \mathcal{L}_{\rm sdf2gs} = |D_{\rm gs} - \textbf{sg}[D_{\rm sdf}]| + (1 - <\textbf{n}_{\rm gs}, \textbf{sg}[\textbf{n}_{\rm sdf}]>).
\end{equation}
Finally, we obtain the mutual supervision $\mathcal{L}_{\rm gs2sdf}$ and $\mathcal{L}_{\rm sdf2gs}$, which bridges the SDF and Gaussian. Note that the stop-gradient operator is used to differ the strength of the supervision of both sides. 

\paragraph{Discussion.}
In general, SDF is optimized via color supervision between volume-rendered images and input images. However, our loss $\mathcal{L}_{\rm gs2sdf}$ can optimize the SDF via depth and normal without color supervision, as the depth and the normal are derived from SDF, and thus the gradient from $\mathcal{L}_{\rm gs2sdf}$ can optimize the SDF. In this way, we can drop the heavy color rendering and optimize the SDF with only geometry attributes (\ie depth and normal). With other widely used regularizations for SDF (\eg Eikonal loss), we can obtain a smooth and reasonable geometry from SDF. The loss $\mathcal{L}_{\rm sdf2gs}$ regularizes the geometry of Gaussians in turn, so we can obtain a better decomposition between geometry and material from the Gaussians.

\subsection{SDF-aware pruning}
\label{sec:prune}

\begin{figure}[tb]
    \centering
    \includegraphics[width = \linewidth]{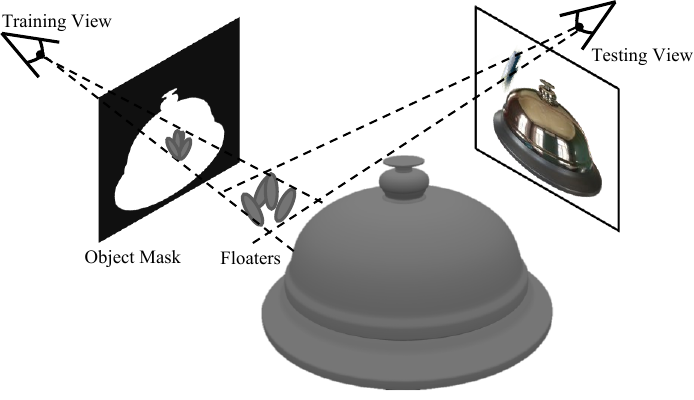}
    \caption{In this example, the floaters can still be shown in the testing view, even if the mask loss was applied during training. The main reason is that although the floaters are shown outside the mask region in the testing view, they are within the mask region in the training views. Therefore, they can not be masked out by the mask loss.}
    \Description[Floaters still exist after the mask loss was applied.]{Some floaters lays in the mask region in the training views, so the mask loss cannot diminish these floaters.}
    \label{fig:floater}
\end{figure}

\begin{figure}[tb]
    \centering
    \includegraphics[width = \linewidth]{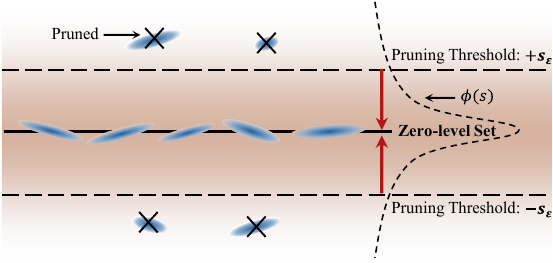}
    \caption{The illustration of SDF-aware pruning. We define a narrowing threshold, which is adjusted automatically around the zero-level set. The Gaussians out of the threshold will be pruned. This pruning operation ensures all Gaussians are near the surface and avoids the floaters.}
    \Description[Illustration of the SDF-aware pruning.]{We compute a threshold from the probability density distribution of SDF and prune the Gaussians out of the threshold.}
    \label{fig:gs_prune}
\end{figure}

With the deferred Gaussians supervised by SDF priors, the geometry becomes smoother. However, we observe some artifacts caused by the overfitting on specular highlights of reflective objects in training views, which leads to Gaussian outliers distant from the surface. These overfitted Gaussians are in the mask region of all training views, so applying the mask loss cannot remove such outliers in training, and we still observe the floaters in the test view, as shown in ~\figref{fig:floater}. Besides, all the above losses supervise the alpha-blended results and lack constraints on individual Gaussians to handle the outliers. Therefore, we propose a Gaussian pruning strategy to enforce the Gaussians close to the zero-level set of SDF.
Specifically, for each Gaussian, we check the SDF value at its center and discard the Gaussian if the SDF value is larger than a threshold $s_\varepsilon$, as shown in the \figref{fig:gs_prune}. 
Since it is laborious to adjust the threshold per scene, we link it with the probability density function $\phi_s(s)$ in \eqref{eq:dpf}, and the threshold $s_\varepsilon$ can be defined as $\phi_s(s_\varepsilon)=p_t (s_\varepsilon > 0)$ in the closed form: 
\begin{equation}
    s_\varepsilon = \gamma\log{(2p_t)} - \gamma\log{\left( \gamma^{-1} - 2p_t - \sqrt{\gamma^{-2}-4p_t\gamma^{-1}}\right)},
\end{equation}

where $p_t$ is an empirical hyperparameter and set to $0.01$.
In this way, the threshold is determined automatically and narrowed with the SDF in the optimization, so we avoid posing a constant too strict when the surface is under-reconstructed and prune the Gaussians too far away from the surface in the end.

\subsection{GS-guided SDF enhancement}
\label{sec:ft}

\begin{figure}[tb]
    \centering
    \includegraphics[width = \linewidth]{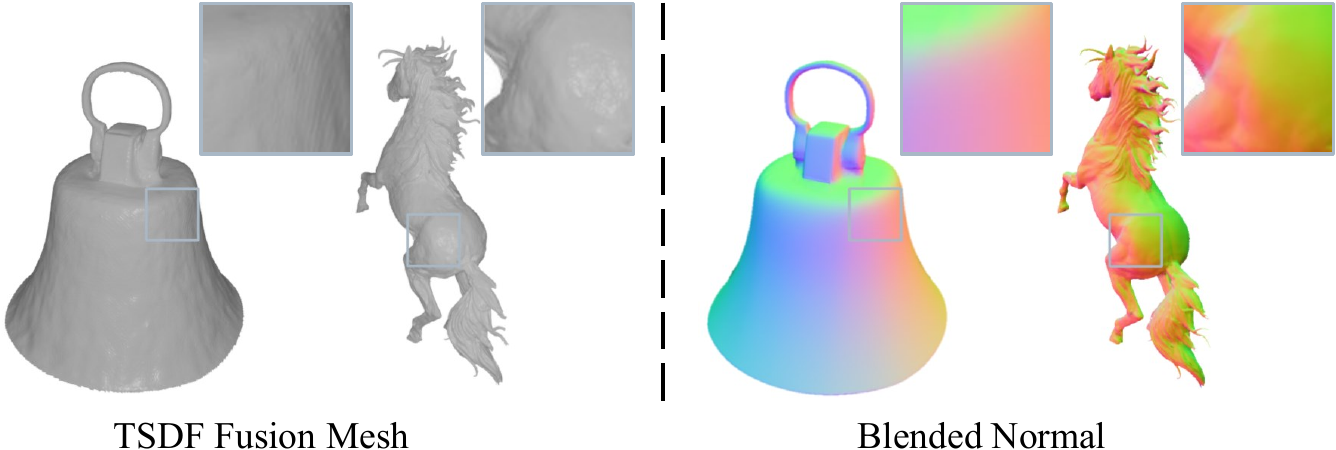}
    \caption{Although the relighting framework ensures smooth blended normal to render realistic glossy highlight, the blended depth is inconsistent with the normal, and thus the mesh of TSDF fusion from depth can be problematic.}
    \Description[Inconsistency between Gaussian depth and normal.]{The predicted normal from the Gaussian model is smooth, while the mesh of TSDF fusion from depth is unsmooth.}
    \label{fig:geo_inconsist}
\end{figure}

\begin{figure}[tb]
    \centering
    \includegraphics[width = \linewidth]{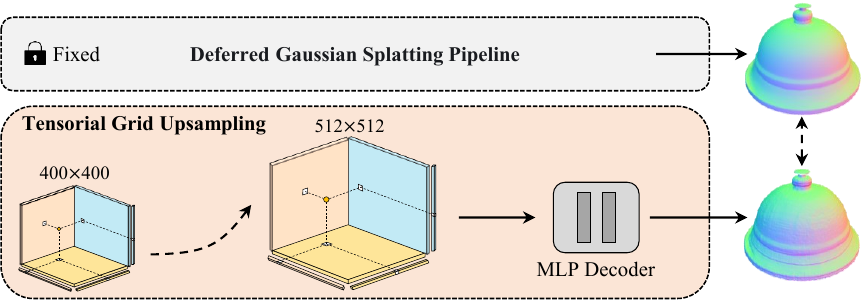}
    \caption{The overview of GS-guided SDF enhancement. The TensoSDF learned from the first step is upsampled and refined by the blended normal from fixed Gaussians.}
    \Description[The overview of GS-guided SDF enhancement.]{The upsampled TensoSDF is supervised by the normal from the fixed deferred Gaussian splatting pipeline.}
    \label{fig:overview_ft}
\end{figure}

Due to the inconsistency between the normal and depth from blended Gaussians, as shown in \figref{fig:geo_inconsist}, the meshes of TSDF fusion from depth can be problematic, even though the normal is well optimized in the GS framework. Lots of works~\cite{yu_2024_gof, lukas_2024_spop, chen_2024_pgsr, loccoz_2024_3dgrt} try to alleviate this inconsistency, while these solutions complicate the rendering pipeline and thus reduce the GS efficiency in training and rendering.

Since our framework has the TensoSDF as an extra geometry representation, we can directly extract mesh from it, without complicating the GS pipeline. Therefore, we propose the GS-guided SDF enhancement, which utilizes the normal from well-optimized Gaussians to enable extracting high-quality meshes from TensoSDF.

Specifically, we first upsample the TensoSDF to capture the fine-grained details, which keeps a low resolution for efficiency in the first step. Then, the upsampled TensoSDF is finetuned efficiently via the learned normal from fixed Gaussians, as shown in \figref{fig:overview_ft}. The loss for finetuning $\mathcal{L}_{\rm ft}$ is
\begin{equation}
    \mathcal{L}_{\rm ft} = (1 - <\textbf{sg}[\textbf{n}_{\rm gs}], \textbf{n}_{\rm sdf}>) + \lambda_{\rm hes}\mathcal{L}_{\rm hes} + \lambda_{m}\mathcal{L}_{\rm m} + \lambda_{\rm tv}\mathcal{L}_{\rm tv},
\end{equation}
where all losses $\mathcal{L}_{[\cdot]}$ share the same definition as in \eqref{eq:l_sdf}.

This way, the TensoSDF learns geometry details from the well-optimized Gaussians. Therefore, our method achieves high-quality geometry reconstruction for reflective objects, which beats the TensoSDF optimized via ray marching and uses only 33\% training time of it. Compared to optimizing a full-resolution SDF and Gaussians jointly as in GSDF, we modulate the resolution of TensoSDF adaptively, \revised{leading to a more efficient framework}. Consequently, our framework maintains the highly efficient training/rendering from Gaussian splatting and meanwhile benefits from the robustness of the TensoSDF representation for geometry reconstruction.

\section{Implementation details}

This section presents the network structures and the training details.

\paragraph{Network structures.} The TensoSDF has a resolution of $400\times 400$ with feature channels set as $36$, which upsamples to $512\times512$ further. The MLP decoder is a two-layer MLP whose width is 128. The Gaussian starts with 100K randomly initialized points.
\begin{table}[tb]
    \centering
    \caption{Losses used in our paper.}
    \resizebox{0.99\linewidth}{!}{
    \begin{tabular}{l|c|c}
    \hline
    Name & Apply on & Description \\
    \hline
       $L_{\rm c}$        & GS     & Color supervision for 3DGS \\
       $L_{\rm nd}$       & GS     & Normal Consistency loss from GShader. \\
       $L_{\rm sm}$       & GS     & BRDF smoothness loss from R3DG. \\
       $L_{\rm \delta n}$ & GS     & Normal regularization for GShader. \\
       $L_{\rm sdf2gs}$   & GS     & The proposed depth/normal supervision for GS. \\
       $L_{\rm gs2sdf}$   & SDF    & The proposed depth/normal supervision for SDF. \\
       $L_{\rm eik}$      & SDF    & Eikonal loss from Gropp et al.~\shortcite{gropp_2020_eik}. \\
       $L_{\rm hes}$      & SDF    & Hession Loss from Zhang et al.~\shortcite{zhang_2022_hes}. \\
       $L_{\rm tv}$       & SDF    & Total variance loss from TensoRF. \\
       $L_{\rm m}$        & GS/SDF & Mask loss for both 3DGS and SDF. \\
    \hline
       
    \end{tabular}
    }
    
    \label{tab:loss_des}
\end{table}

\begin{table*}[t]
    \centering
    \caption{Relighting quality in terms of PSNR$\uparrow$ and SSIM$\uparrow$ on the Glossy Blender dataset. Numbers in \sota{red} indicate the best performance, \subsota{orange} numbers indicate the second best, and numbers in \third{yellow} indicate the third best. Our method outperforms existing Gaussian-based methods. Note that although our quantitative metrics are lower than some NeRF-based methods, our training time is much shorter (25\% of TensoSDF, 13\% of NeRO).}
\resizebox{\linewidth}{!}{
    \renewcommand{\arraystretch}{1.1}
    \begin{tabular}{l|cccc|cccc}
    \hline
            & \multicolumn{4}{c|}{NeRF-based} & \multicolumn{4}{c}{Gaussian-based}   \\ 
            & MII & TensoIR & TensoSDF & NeRO & GShader & GS-IR & R3DG & Ours  \\
            & PSNR/SSIM & PSNR/SSIM & PSNR/SSIM & PSNR/SSIM & PSNR/SSIM & PSNR/SSIM & PSNR/SSIM & PSNR/SSIM  \\
       \hline
       Angel  & 16.24/.8236 & 10.24/.2238 & \subsota{20.40/.8969} & 16.21/.7819 & \third{17.49/.8336} & 15.64/.6126 & 16.65/.8013 & \sota{20.81/.8775} \\
       Bell   & 17.41/.8594 & 10.11/.1018 & \subsota{29.91/.9767} & \sota{31.19/.9794} & 19.01/.8804 & 12.61/.2807 & 16.15/.8391 & \third{24.49/.9267} \\
       Cat    & 17.68/.8521 & 9.10/.1644 & \third{26.12/.9354} & \sota{28.42/.9579} & 16.00/.8642 & 18.04/.7907 & 17.49/.8503 & \subsota{26.28/.9421} \\ 
       Horse  & 20.98/.8997 & 10.42/.1931 & \sota{27.18/.9567} & \subsota{25.56/.9437} & 22.49/.9262 & 17.40/.7270 & 20.63/.8832 & \third{23.31/.9376} \\
       Luyu   & 17.89/.8050 & 8.27/.2375 & \third{19.91/.8825} & \sota{26.22/.9092} & 15.62/.8254 & 19.00/.7727 & 17.47/.8168 & \subsota{22.61/.8995} \\ 
       Potion & 17.13/.8094 & 6.21/.0846 & \subsota{27.71/.9422} & \sota{30.14/.9561} & 12.33/.7575 & 16.37/.7051 & 14.99/.7799 & \third{25.67/.9175} \\
       Tbell  & 16.54/.8262 & 7.47/.1609 & \subsota{23.33/.9404} & \sota{25.45/.9607} & 14.42/.8007 & 14.35/.5419 & 15.99/.7965 & \third{22.80/.9180} \\
       Teapot & 16.71/.8546 & 9.96/.2093 & \subsota{25.16/.9482} & \sota{29.87/.9755} & 18.21/.8560 & 16.63/.7646 & 17.36/.8389 & \third{21.17/.8932} \\
       \hline
       Mean & 17.57/.8413 & 8.97/.1719 & \subsota{24.97/.9349} & \sota{26.63/.9331} & 16.95/.8430 & 16.26/.6494 & 17.09/.8258 & \third{23.39/.9140}  \\
       \hline
       Training Time  & 4h & 5h &  6h & 12h & 0.5h & 0.5h & 1h  &    1.5h        \\
       Ren. Time (FPS)   & 1/30 & 1/60 &  1/4 & 1/4 & 50 & 214  & 1.5   &    208        \\
       \hline
        
    \end{tabular}
}
    
    \label{tab:nero_syn}
\end{table*}

\paragraph{Optimization details}
\label{sec:opt}
The training of our framework includes three steps. Firstly, the deferred Gaussian splatting without SDF supervision is trained separately for 1K iterations with \eqref{eq:l_gs} to obtain a coarse geometry. Then, we freeze the Gaussians and warm up the TensoSDF via the depth and normal from Gaussians \eqref{eq:l_sdf} for 3K iterations. Finally, we train the entire framework with a joint loss for 24K iterations: 
\begin{equation}
    \mathcal{L}_{\rm joint} = \mathcal{L}_{\rm gs} + \lambda_{\rm sdf2gs}\mathcal{L}_{\rm sdf2gs}.
    \label{eq:ljoint}
\end{equation}
We provide a table of loss functions (see ~\tabref{tab:loss_des}) for better understanding and more details are in the supplementary material.  

We train our model on an NVIDIA RTX 4090 and optimize it with Adam optimizer~\cite{kingma_2014_adam}. The learning rate for the TensoSDF~\cite{li_2024_tensosdf} grid and the MLP decoder is $1e^{-2}$ and $1e^{-3}$, respectively. The learning rate of Gaussian components follows the setting in 3DGS~\cite{kerbl_2023_3dgs}, and all attributes not in vanilla 3DGS have a learning rate of $2e^{-3}$. The initial resolution of TensoSDF is set to $128\time 128$ and increases to the final resolution $400\times 400$, which uses only 60\% capacity of the original TensoSDF, whose final resolution is $512\times 512$. The optimization procedure takes about 1.5 hours in total.
After training, only Gaussians will be used for relighting. To extract high-quality meshes from TensoSDF, a 15-minute finetuning is needed, using the Adam optimizer and a cosine learning rate schedule. The learning rate for the TensoSDF~\cite{li_2024_tensosdf} grid and the MLP decoder starts with $1e^{-2}$ and $1e^{-3}$, which decays to $5e^{-4}$ and $5e^{-5}$ finally.

%

\section{Results}
\label{sec:results}
In this section, we describe the evaluation setup (\secref{sec:eval_setup}), and evaluate our relighting and NVS results (\secref{sec:res}). Then, we conduct ablation studies on several main components (\secref{sec:ab}) and discuss limitations (\secref{sec:lim}). More results and comparison can be found in the supplementary file and the demo video.

\subsection{Evaluation setup}
\label{sec:eval_setup}

\paragraph{Dataset} We evaluate our method on three synthetic datasets and one real-scene dataset. We utilize the TensoIR~\cite{jin_2023_tensoir} synthetic and Glossy Blender dataset~\cite{liu_2023_nero} to evaluate the performance of diffuse and specular materials, respectively. Besides, we use the Shiny Blender dataset to show the generalization ability of our method for diverse materials. For real scenes, we select some specular objects from the NeILF++\cite{zhang_2023_neilfpp} and \revised{some representative ones from the Stanford-ORB dataset~\cite{kuang_2023_stanfordorb} in the supplementary materials.}

\paragraph{Methods for comparison}  We select 7 representative methods for comparison. We choose MII~\cite{zhang_2022_mii}, TensoIR~\cite{jin_2023_tensoir}, TensoSDF~\cite{li_2024_tensosdf}, and NeRO~\cite{liu_2023_nero} for NeRF-based methods and GShader~\cite{jiang_2024_gaussianshader}, GS-IR~\cite{liang_2024_gsir}, and R3DG~\cite{gao_2023_relightablegs} for Gaussian-based methods. While evaluating the surface quality, we also select some non-relightable Gaussian-based reconstruction methods, including 2DGS, GOF~\cite{yu_2024_gof}, and PGSR~\cite{chen_2024_pgsr}. We trained these models based on their public codes and configurations. Note that we add the mask loss to train all the compared methods, for stabilizing training and fair comparison.

\paragraph{Metrics} We use the peak signal-to-noise ratio (PSNR) and structural similarity index (SSIM)~\cite{wang_2004_ssim} to measure the relighting quality of results for the comparison and ablation study in the main paper. We present the learned perceptual image patch similarity (LPIPS)~\cite{zhang_2018_lpips} in the supplementary materials. Due to the ambiguity between albedo and light, we normalize the relighting images to match the average colors of ground-truth images before computing the metrics, following NeRO.
When the ground truth albedo is available, we rescale the relighting images based on the ground truth albedo and predicted albedo, following TensoIR. We use the mean angular error (MAE) and the Chamber distance (CD) to measure the quality of geometry reconstruction. We record the mean training time and the frame per second (FPS)\footnote{NeRO and TensoSDF use the Cycles Render Engine in Blender to provide relighting results, so the FPS depends on samples per pixel, which is 1024 following the NeRO setting.} in the evaluation. All the FPS and training time are tested on an RTX 4090 unless otherwise specified.

\subsection{Comparison with previous works}
\label{sec:res}

\begin{figure*}[htb]
    \centering
    \includegraphics*[clip, width = 0.98\linewidth]{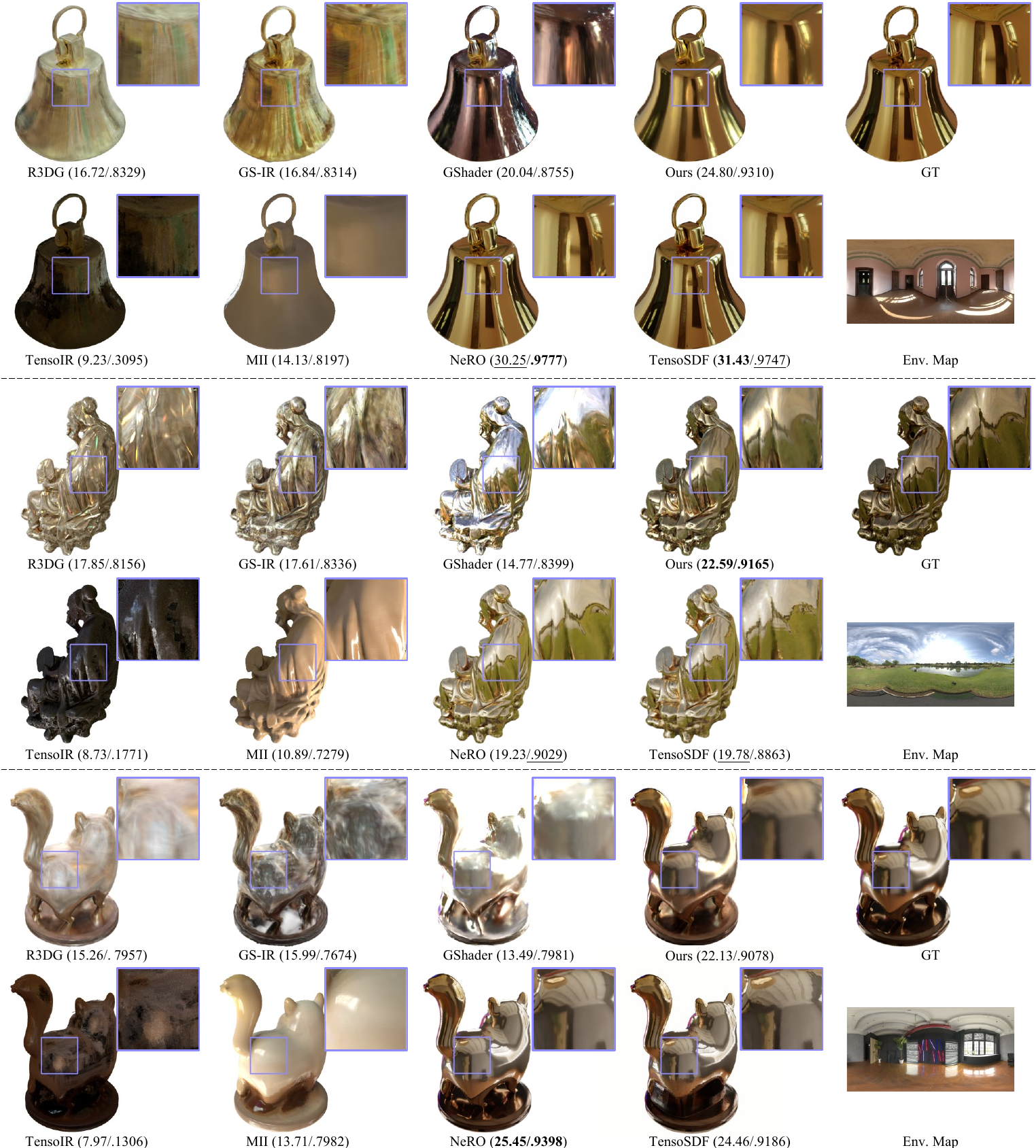}
    \caption{Relighting comparison on the Glossy Blender Dataset. R3DG, GS-IR, TensoIR, and MII fail to model specular highlights. GShader provides specular relighting results with obvious artifacts. Our method, along with NeRO and TensoSDF, provides high-quality results, while our method only uses 25\% of the training time of TensoSDF and enables real-time relighting. We present the metrics per scene (PSNR/SSIM).}
    \Description[Relighting results on the Glossy Blender Dataset]{We show three cases of reflective objects on the Glossy Blender dataset. Our method can produce relighting results with faithful highlights while Gaussian-based methods fail. However, the results from our method are more blurry than the ones from NeRO and TensoSDF.}
    \label{fig:relit_glossy}
\end{figure*}

\begin{figure*}[htb]
    \centering
    \includegraphics*[clip, width = 0.98\linewidth]{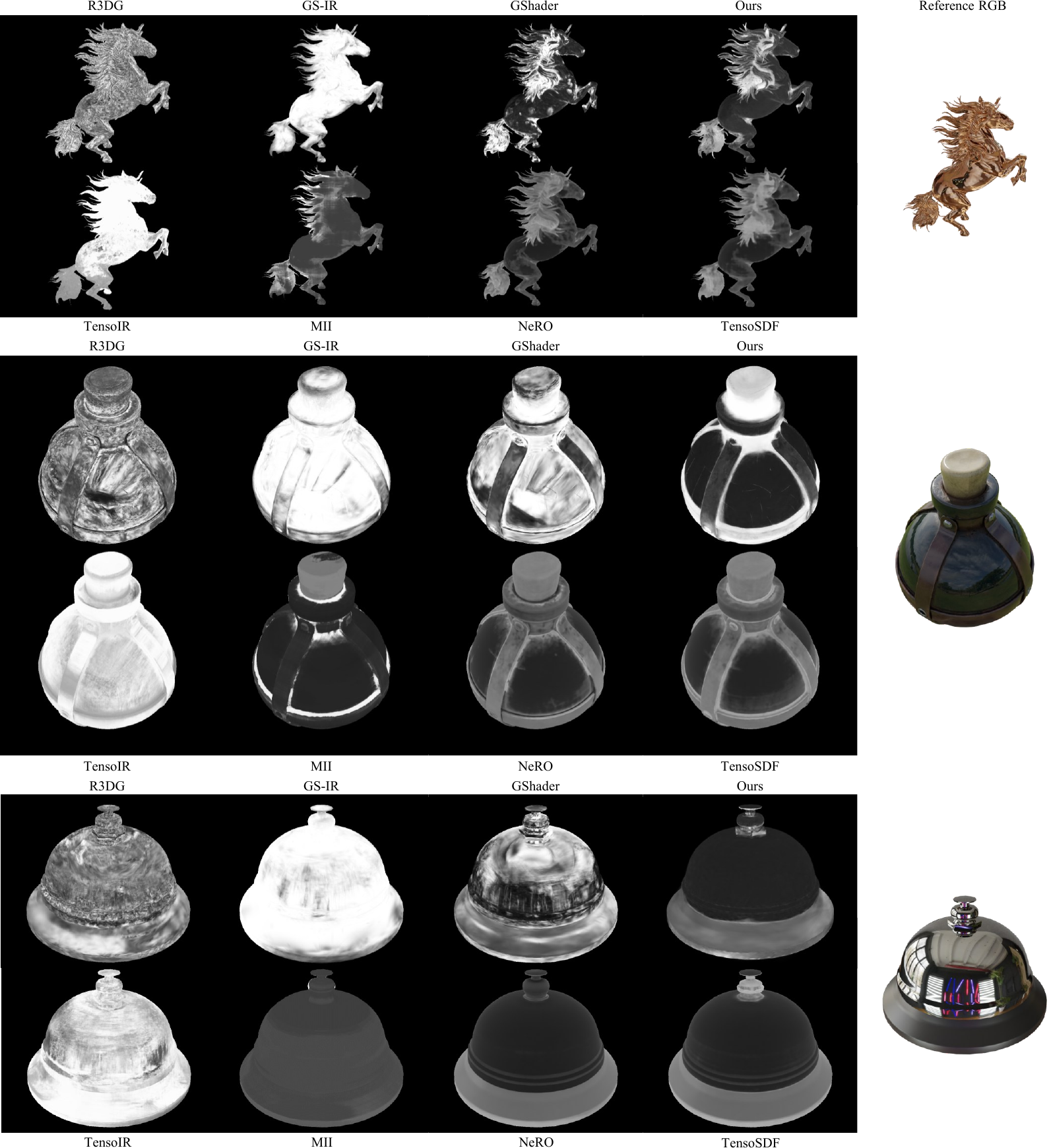}
    \caption{Decomposed roughness comparison on the Glossy Blender Dataset. Gaussian-based methods and TensoIR reconstruct problematic roughness. MII provides over-smooth prediction and ignores the difference of material. Our method, NeRO, and TensoSDF reconstruct reasonable roughness for relighting, while our method takes 25\% training time at most.}
    \Description[Decomposed roughness on the Glossy Blender Dataset]{We show three cases of reflective objects on the Glossy Blender dataset. Our method and NeRF-based methods can produce reasonable decomposed roughness while Gaussian-based methods fail.}
    \label{fig:roughness_glossy}
\end{figure*}

\begin{figure*}[htb]
    \centering
    \includegraphics*[clip, width = \linewidth]{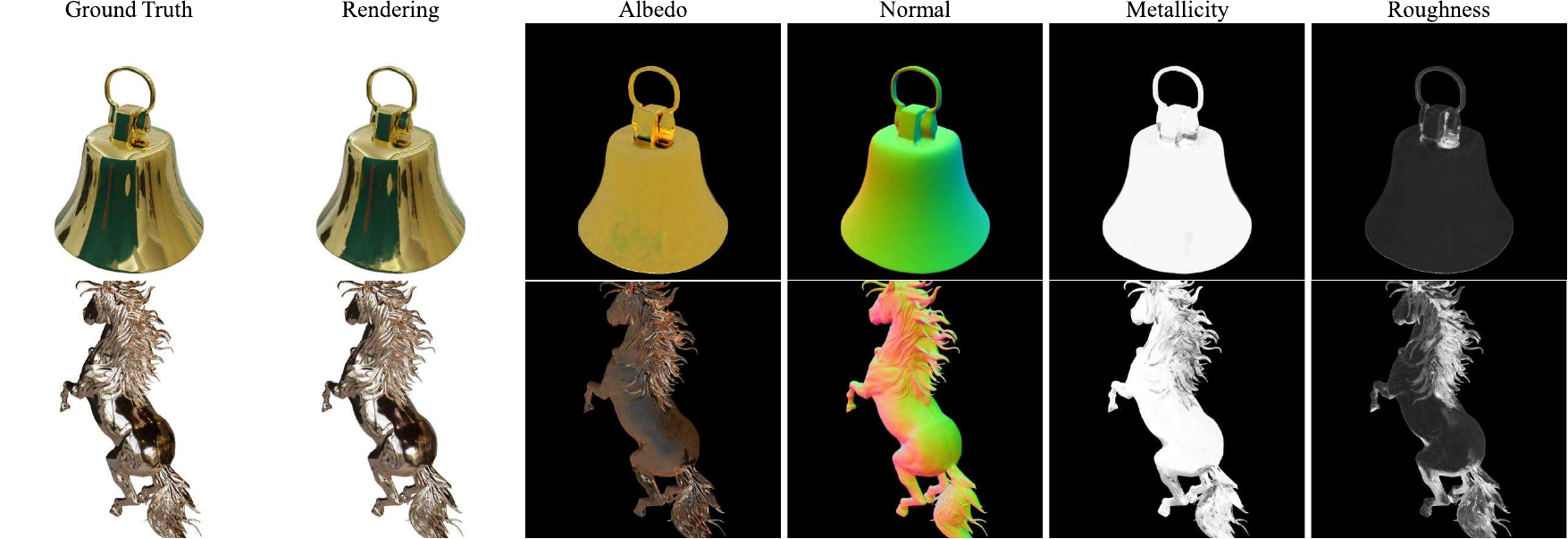}
    \caption{Decomposed maps of our method on the Glossy Blender Dataset. Our method can provide a reasonable decomposition for reflective surfaces. The normal is smooth but maintains details (see the horse mane in the 2nd row). There is no material ground truth, so we only provide the qualitative visualization.}
    \Description[Decomposed maps on the Glossy Blender Dataset]{This figure shows more decomposed maps from our method, including a bell and a horse. The decomposed maps are continuous and reasonable.}
    \label{fig:decomp}
\end{figure*}

\begin{figure*}[htb]
    \centering
    \includegraphics*[clip, width = \linewidth]{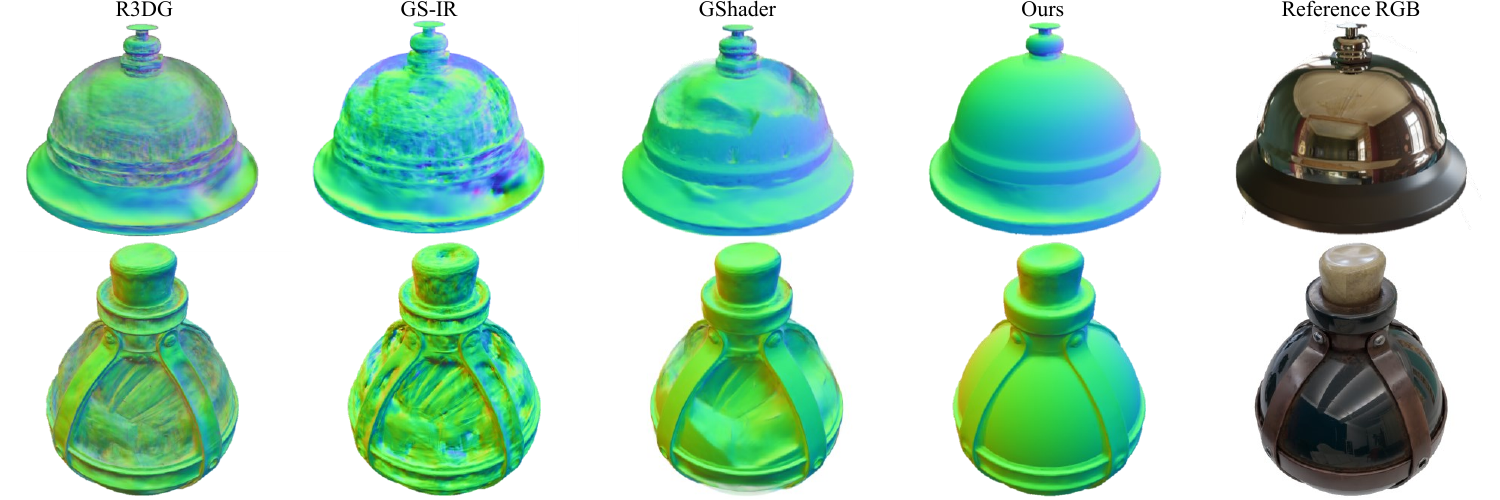}
    \caption{Normal comparison among Gaussian-based methods on the Glossy Blender Dataset. Our method can reconstruct high-quality normals for relighting, while others overfit under the training scene and provide erroneous normals.}
    \Description[Normal comparison on the Glossy Blender Dataset]{Our method provides smooth surfaces while the surfaces from other Gaussian-based methods are broken.}
    \label{fig:normal_cmp_glossy}
\end{figure*}

\paragraph{Relighting of reflective objects}
We evaluate the relighting performance of specular materials on the Glossy Blender dataset. The quantitative measurements are shown in \tabref{tab:nero_syn}, and the relighting visualizations are placed in \figref{fig:relit_glossy}. Our method outperforms other Gaussian-based methods and achieves competitive results of the NeRF-based methods. We also show the decomposed maps in \figref{fig:roughness_glossy}, \figref{fig:decomp}, and \figref{fig:normal_cmp_glossy}. Due to the SDF priors and the deferred Gaussian splatting pipeline, our method can produce a smooth surface without losing details and predict reasonable BRDF parameters.

\begin{figure*}[htb]
    \centering
    \includegraphics*[clip, width = \linewidth]{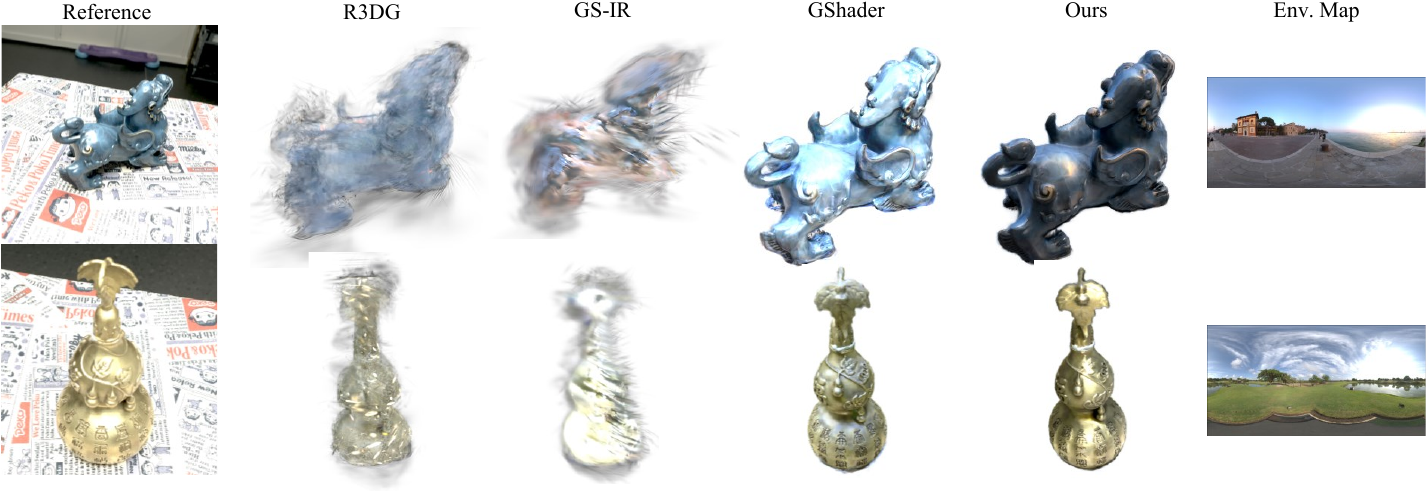}
    \caption{Relighting results on real data from NeILF++. R3DG and GS-IR fail to reconstruct the reflective objects in real scenes. GShader cannot relight the specular highlights faithfully, while our method can provide reasonable relighting results.}
    \Description[Relighting results on real data from NeILF++]{Our method provides realistic relighting results on real data.}
    \label{fig:relit_neilf}
\end{figure*}

\paragraph{Relighting of diffuse objects}
Although our design focuses on reflective surfaces, it can also show benefits for diffuse objects. We validate our method on the TensoIR dataset. The quantitative measurements are shown in \tabref{tab:tir_syn}. The comparisons with NeRF-based methods and relighting visualizations are in the supplementary material. Our methods can outperform the existing Gaussian-based methods and some NeRF-based methods on the TensoIR dataset, which reveals the robustness under diverse materials.

\paragraph{Relighting on real data}
For real scenes, we select some objects with specular highlights from NeILF++~\cite{zhang_2023_neilfpp}, and the results are shown in \figref{fig:relit_neilf}. As there is no available ground truth, we only show the reference training views and some results under novel light conditions. Our method produces reasonable relighting results under novel lighting. Besides, our method achieves a balance between quality and training speed, enabling reflective surface modeling with only 25\% training time of TensoSDF and 13\% of NeRO. Our method supports real-time rendering with 200+ FPS, much faster than all NeRF-based and some Gaussian-based methods.

\begin{table}[tb]
    \centering
    \caption{Relighting quality in terms of PSNR$\uparrow$  and SSIM$\uparrow$ on the TensoIR synthetic dataset. Numbers in \sota{red} indicate the best performance, and numbers in \subsota{orange}  indicate the second best.}
\resizebox{0.99\linewidth}{!}{
    \renewcommand{\arraystretch}{1.3}
    \begin{tabular}{lcccccc}
       \hline
            & GShader & GS-IR & R3DG & Ours  \\
            & PSNR/SSIM & PSNR/SSIM & PSNR/SSIM & PSNR/SSIM  \\
       \hline
       Armad.    & 22.86/.9280 & 27.65/.9078 & \subsota{30.76/.9526} & \sota{31.33/.9593} \\
       Ficus     & 24.61/.9491 & 23.63/.8662 & \sota{27.23/.9637} & \subsota{26.28/.9542} \\
       Hotdog    & 17.45/.8838 & 21.51/.8853 & \subsota{24.59/.9162} & \sota{25.21/.9307} \\ 
       Lego      & 13.41/.7904 & \subsota{22.88/.8342} & 22.49/.8682 & \sota{25.46/.9083} \\
       \hline
       Mean      & 19.58/.8878 & 23.92/.8734 & \subsota{26.27/.9252} & \sota{27.07/.9381} \\
       \hline
    \end{tabular}
    }
    \label{tab:tir_syn}
\end{table}

\begin{table}[tb]
    \centering
    \caption{NVS quality in terms of PSNR$\uparrow$ and SSIM$\uparrow$ on the Glossy Blender dataset. Numbers in \sota{red} indicate the best performance, and numbers in \subsota{orange}  indicate the second best.}
\resizebox{0.99\linewidth}{!}{
    \renewcommand{\arraystretch}{1.2}
    \begin{tabular}{lcccccc}
       \hline
            & GS-IR & R3DG & GShader & Ours  \\
            & PSNR/SSIM & PSNR/SSIM & PSNR/SSIM & PSNR/SSIM  \\
            
       \hline
       Angel   & 20.77/.6935 & 23.12/.7841 & \subsota{27.14/.9226} & \sota{29.32/.9445} \\
       Bell    & 19.05/.4895 & 24.38/.8952 & \subsota{30.00/.9409} & \sota{31.53/.9694} \\
       Cat     & 27.91/.9075 & 29.73/.9487 & \subsota{31.25/.9604} & \sota{31.72/.9672} \\ 
       Horse   & 20.85/.6835 & 22.88/.7769 & \subsota{26.03/.9314} & \sota{27.09/.9479} \\
       Luyu    & 24.96/.8637 & 25.97/.8986 & \subsota{27.35/.9175} & \sota{28.53/.9383} \\ 
       Potion  & 25.81/.8217 & 29.03/.9277 & \subsota{29.53/.9357} & \sota{30.51/.9503} \\ 
       Tbell   & 20.72/.6193 & 23.03/.8878 & \subsota{23.86/.9031} & \sota{29.48/.9648} \\ 
       Teapot  & 20.56/.8427 & 20.82/.8719 & \subsota{23.56/.8986} & \sota{26.41/.9468} \\ 
       \hline
       Mean      & 22.58/.6322 & 24.87/.8739 & \subsota{27.34/.9263} & \sota{29.32/.9537}  \\
       \hline
    \end{tabular}
    }
    \label{tab:nvs_glossy}
\end{table}

\begin{table}[hb]
    \centering
    \caption{NVS quality with Gaussian-based methods on Shiny Blender dataset in terms of PSNR$\uparrow$ and SSIM$\uparrow$. Numbers in \sota{red} indicate the best performance, and numbers in \subsota{orange} indicate the second best.}
\resizebox{0.99\linewidth}{!}{
    \renewcommand{\arraystretch}{1.2}
    \begin{tabular}{lcccccc}
       \hline
            & GS-IR & R3DG & GShader & Ours  \\
            & PSNR/SSIM & PSNR/SSIM & PSNR/SSIM & PSNR/SSIM  \\
            
       \hline
       Ball     & 18.30/.7584 & 21.39/.9047 & \subsota{30.40/.9623} & \sota{35.50/.9849}\\
       Car      & 25.30/.8867 & 26.59/.9267 & \subsota{28.39/.9388} & \sota{30.52/.9638} \\ 
       Coffee   & 30.72/.9463 & \sota{32.57/.9710} &  \subsota{30.79/.9690} & 29.64/.9568 \\ 
       Helmet   & 25.08/.9018 & 26.95/.9469 & \subsota{28.78/.9549} & \sota{32.62/.9738} \\ 
       Teapot   & 38.21/.9900 & \subsota{43.86/.9963} & 43.35/.9957 & \sota{43.88/.9964} \\ 
       Toaster  & 18.66/.7418 & 20.07/.8745 & \subsota{23.95/.9130} & \sota{25.89/.9379} \\ 
       \hline
       Mean      & 26.05/.8708 & 28.57/.9367 & \subsota{30.94/.9556} & \sota{33.01/.9689} \\
       \hline
    \end{tabular}
    }
    \label{tab:shiny_nvs_psnr}
\end{table}

\begin{table}[hb]
    \centering
    \caption{Normal quality with Gaussian-based methods on Shiny Blender dataset in terms of MAE$\downarrow$. Numbers in \sota{red} indicate the best performance, and numbers in \subsota{orange} indicate the second best.}
    \renewcommand{\arraystretch}{1.0}
    \begin{tabular}{lcccccc}
       \hline
            & GS-IR & R3DG & GShader & Ours  \\
       \hline
       Ball     & 25.79 & 22.44 & \subsota{7.03}  & \sota{0.92} \\
       Car      & 28.31 & 26.02 & \subsota{14.05} & \sota{11.98} \\ 
       Coffee   & 15.38 & \subsota{13.39} & 14.93 & \sota{12.24} \\ 
       Helmet   & 25.58 & 19.63 & \subsota{9.33}  & \sota{4.10}  \\ 
       Teapot   & 15.35 & 9.21  & \subsota{7.17}  & \sota{5.88}  \\ 
       Toaster  & 33.51 & 28.17 & \subsota{13.08} & \sota{8.24}  \\ 
       \hline
       Mean     & 23.99 & 19.81 & \subsota{10.93} & \sota{7.23} \\
       \hline
    \end{tabular}
    
    \label{tab:shiny_mae}
\end{table}

\begin{table}[tb]
    \centering
    \caption{Mesh quality in terms of CD$\downarrow$ on the Glossy Blender dataset. (CD is multiplied by $10^2$) Numbers in \sota{red} indicate the best performance, and numbers in \subsota{orange} indicate the second best. Our method outperforms existing Gaussian-based methods.}
\resizebox{\linewidth}{!}{
    \begin{tabular}{l|ccc|cccc}
    \hline
            & \multicolumn{3}{c|}{Non-relightable} & \multicolumn{4}{c}{Relightable} \\ 
             & 2DGS & GOF & PGSR &  GShader & GS-IR & R3DG & Ours  \\
       \hline
       Angel   & 0.93 & 0.73 & \subsota{0.54} & 0.85 & 1.77 & 0.98 & \sota{0.41} \\
       Bell    & 3.35 & 3.07 & 1.65 & \subsota{1.10} & 11.53 & 4.18 & \sota{0.31} \\
       Cat     & 2.74 & \subsota{2.25} & 2.93 & 2.56 & 5.88 & 3.39 & \sota{1.34} \\ 
       Horse   & 1.18 & 1.07 & \subsota{0.73} & \subsota{0.73} & 1.96 & 1.35 & \sota{0.34} \\
       Luyu    & 1.35 & 1.46 & 0.85 & \subsota{1.07} & 2.25 & 1.68 & \sota{0.81} \\ 
       Potion  & 4.50 & 4.13 & \subsota{1.96} & 4.74 & 6.23 & 3.80 & \sota{0.75} \\
       Tbell   & 5.50 & \subsota{4.61} & 4.72 & 5.74 & 10.21 & 5.00 & \sota{0.55} \\
       Teapot  & 2.18 & 3.32 & \subsota{1.24} & 3.40 & 7.19 & 4.79 & \sota{0.47} \\
       \hline
       Mean   & 2.72 & 2.58 & \subsota{2.44} & 2.53 & 5.88 & 3.15 & \sota{0.62} \\
       \hline
    \end{tabular}
}
    
    \label{tab:nero_cd_gs}
\end{table}


\begin{figure*}[htb]
    \centering
    \includegraphics*[clip, width = \linewidth]{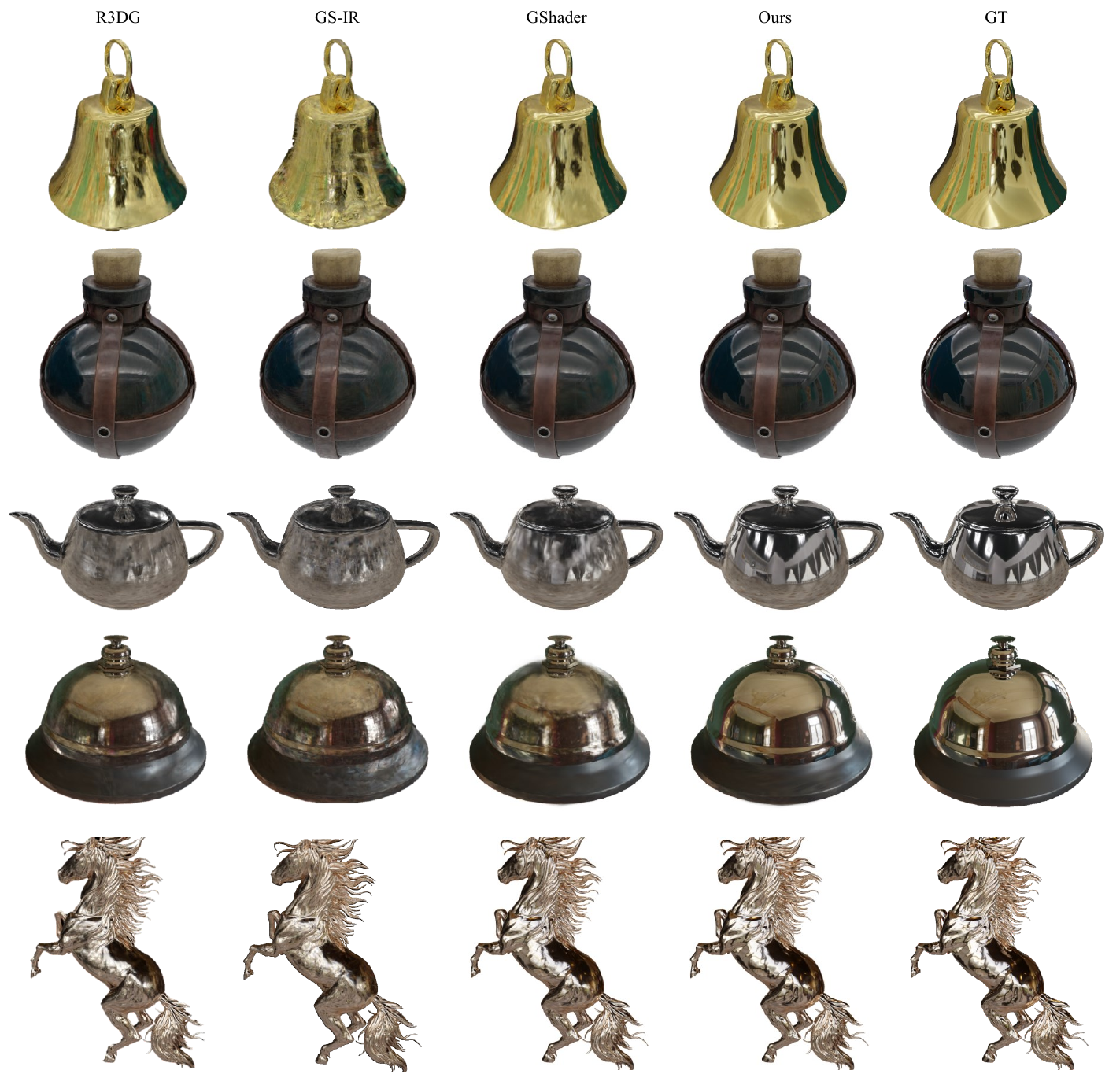}
    \caption{NVS results on the Glossy dataset. Our method provides photo-realistic NVS results for reflective objects, while other Gaussian-based methods fail.}
    \Description[NVS comparison on the Glossy dataset]{Our method provides photo-realistic NVS results for reflective objects, while other Gaussian-based methods cannot reconstruct the specular highlights.}
    \label{fig:nvs_glossy}
\end{figure*}
\begin{figure*}[htb]
    \centering
    \includegraphics*[clip, width = 0.98\linewidth]{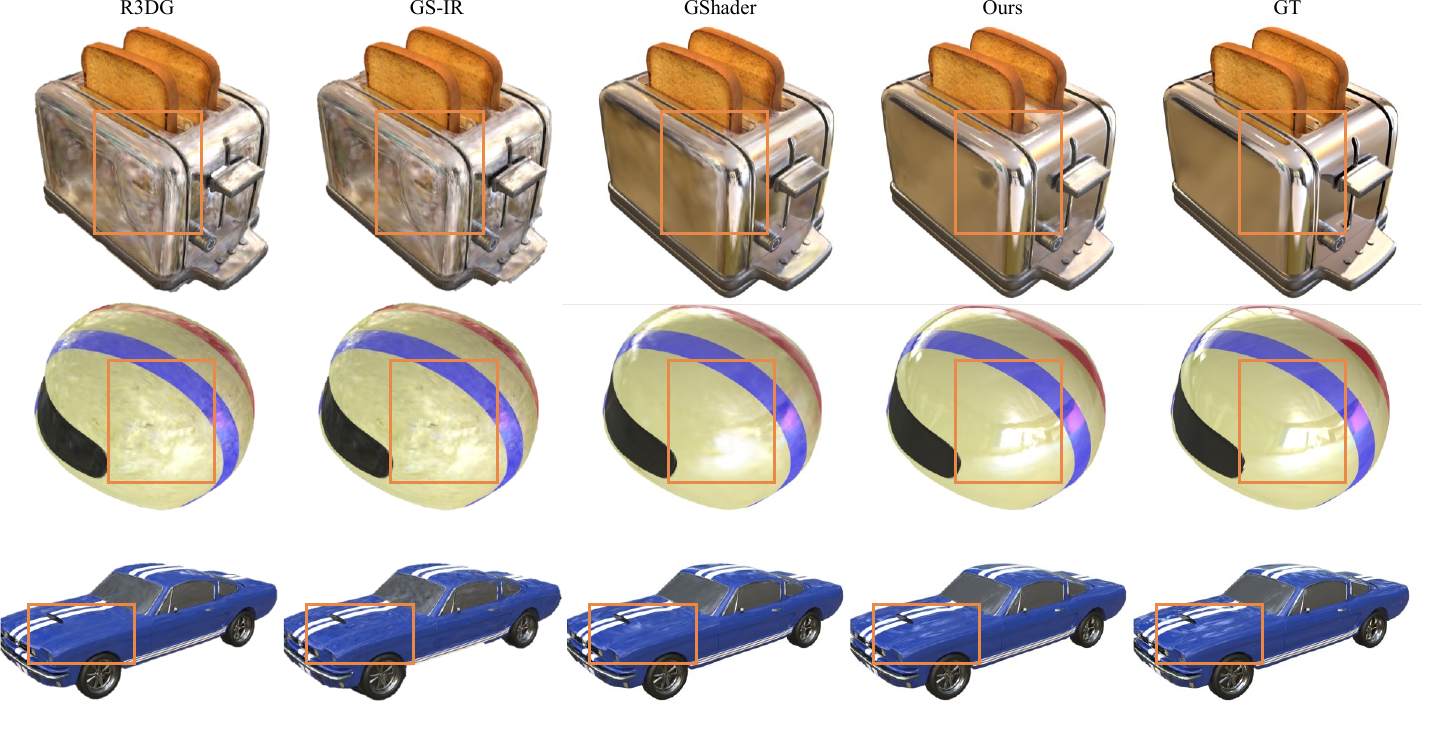}
    \vspace{-2mm}
    \caption{NVS results of Gaussian-based methods on the Shiny Blender dataset. Our method preserves correct highlights, while others fail to render sharp highlights.}
    \vspace{-3mm}
    \Description[NVS comparison on the Shiny Blender dataset]{Our method provides photo-realistic NVS results for reflective objects on the Shiny Blender dataset, while other Gaussian-based methods cannot reconstruct the specular highlights.}
    \label{fig:nvs_shiny}
\end{figure*}
\begin{figure*}[htb]
    \centering
    \includegraphics*[clip, width = 0.98\linewidth]{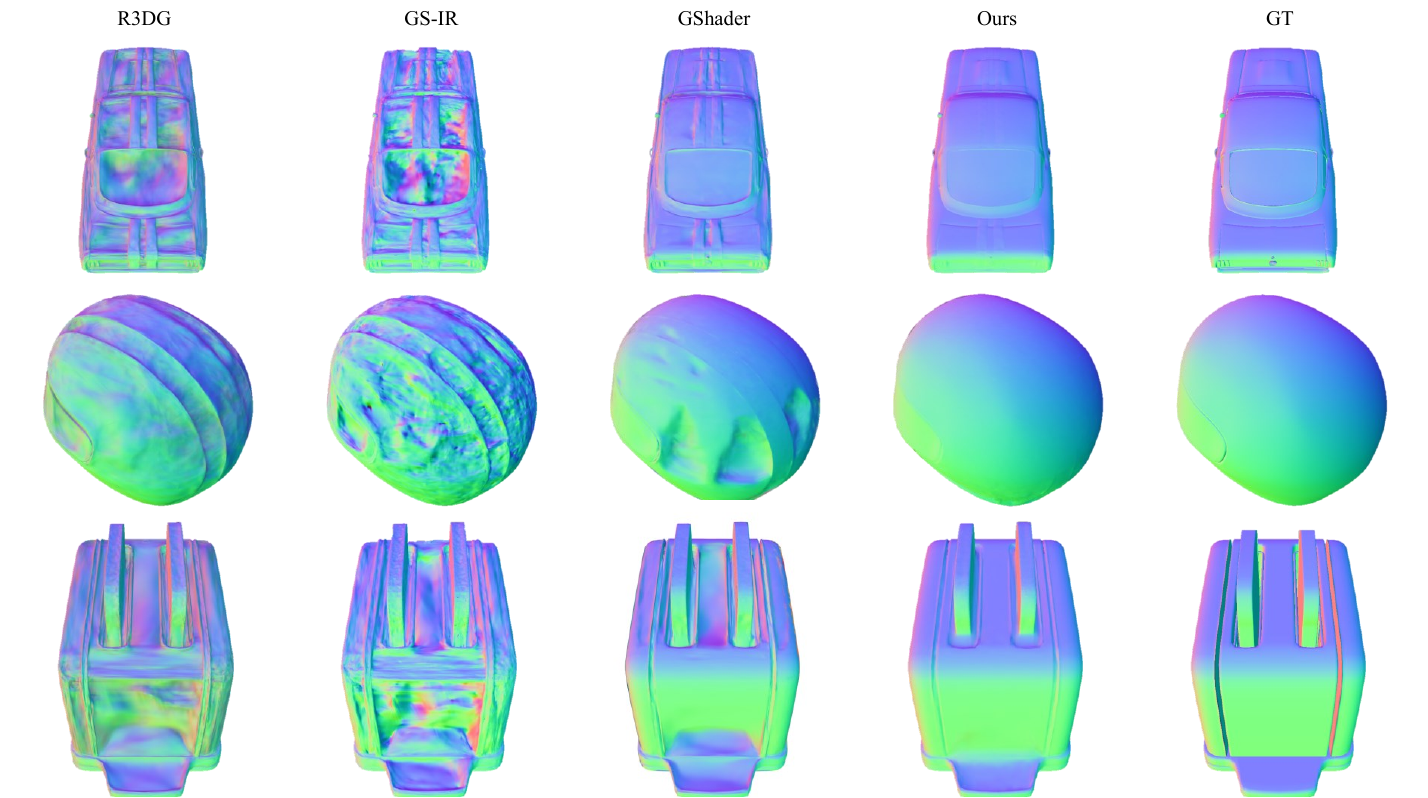}
    \vspace{-2mm}
    \caption{Normal comparison with Gaussian-based methods on the Shiny Blender dataset. Our method provides robust normal estimation, while other results are noisy or overfitted.}
    \Description[Normal comparison on the Shiny Blender dataset]{Our method provides smooth and reasonable reconstructed normals while the normals from other Gaussian-based methods are discontinuous.}
    \label{fig:normal_cmp_shiny}
\end{figure*}
\begin{figure*}[htb]
    \centering
    \includegraphics*[clip, width = 0.98\linewidth]{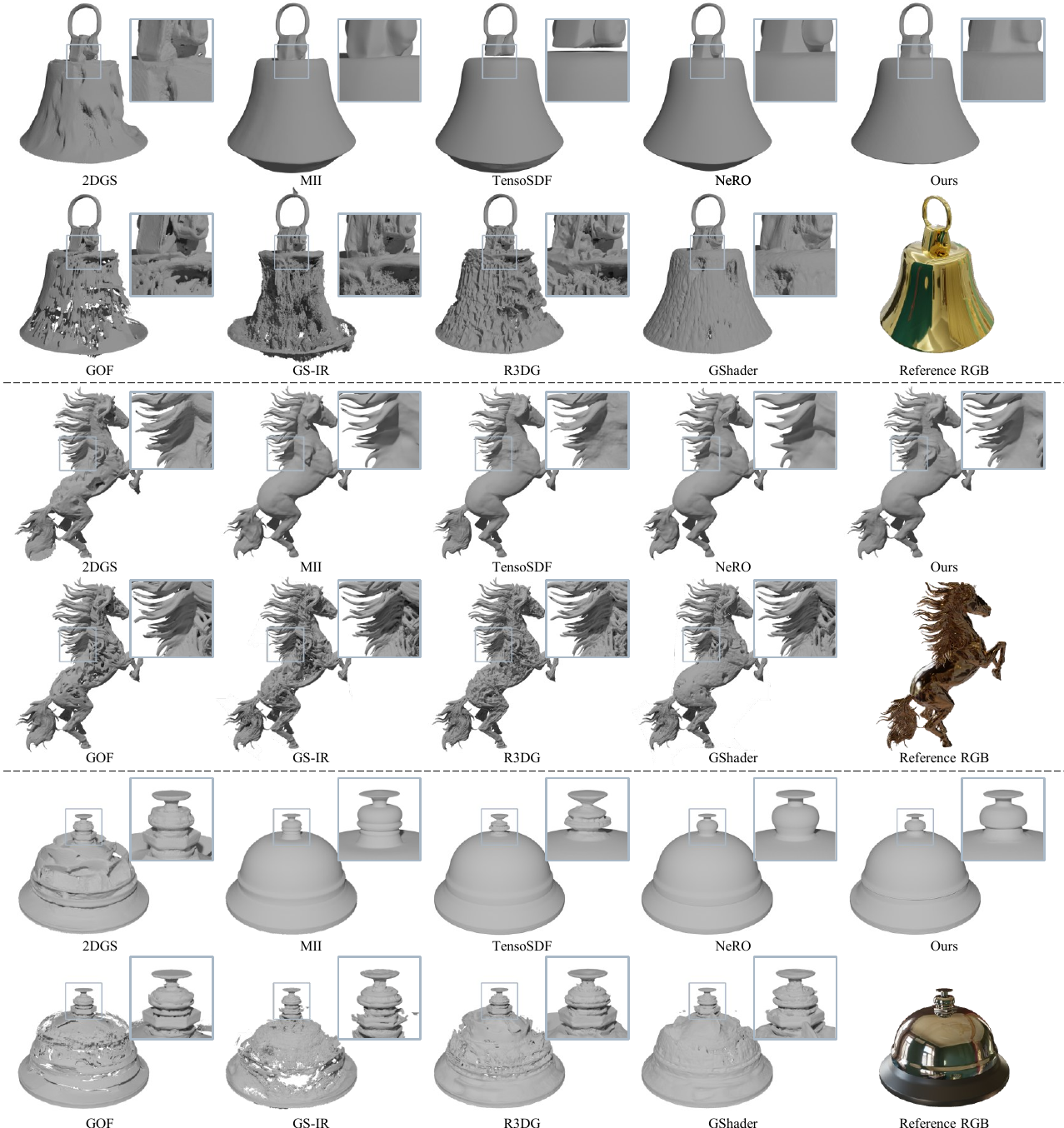}
    \vspace{-1mm}
    \caption{Mesh comparison on the Glossy Blender dataset. Our method provides high-quality meshes from the finetuned TensoSDF. All Gaussian-based methods fail to reconstruct smooth surfaces for reflective objects, while NeRF-based methods cannot preserve the geometric details. Our method ensures both global smoothness and local details. Note that the ground-truth meshes of this dataset are unavailable (only point cloud is available), so we choose to show the RGB reference under the nearest viewpoint on the dataset. Due to the limited space, we present the PGSR results in the supplementary.}
    \Description[Mesh comparison on the Glossy Blender dataset]{Our method provides smooth and reasonable reconstructed meshes while the meshes from other Gaussian-based methods are discontinuous.}
    \label{fig:mesh_glossy}
\end{figure*}

\begin{figure*}[htb]
    \centering
    \includegraphics*[clip, width = \linewidth]{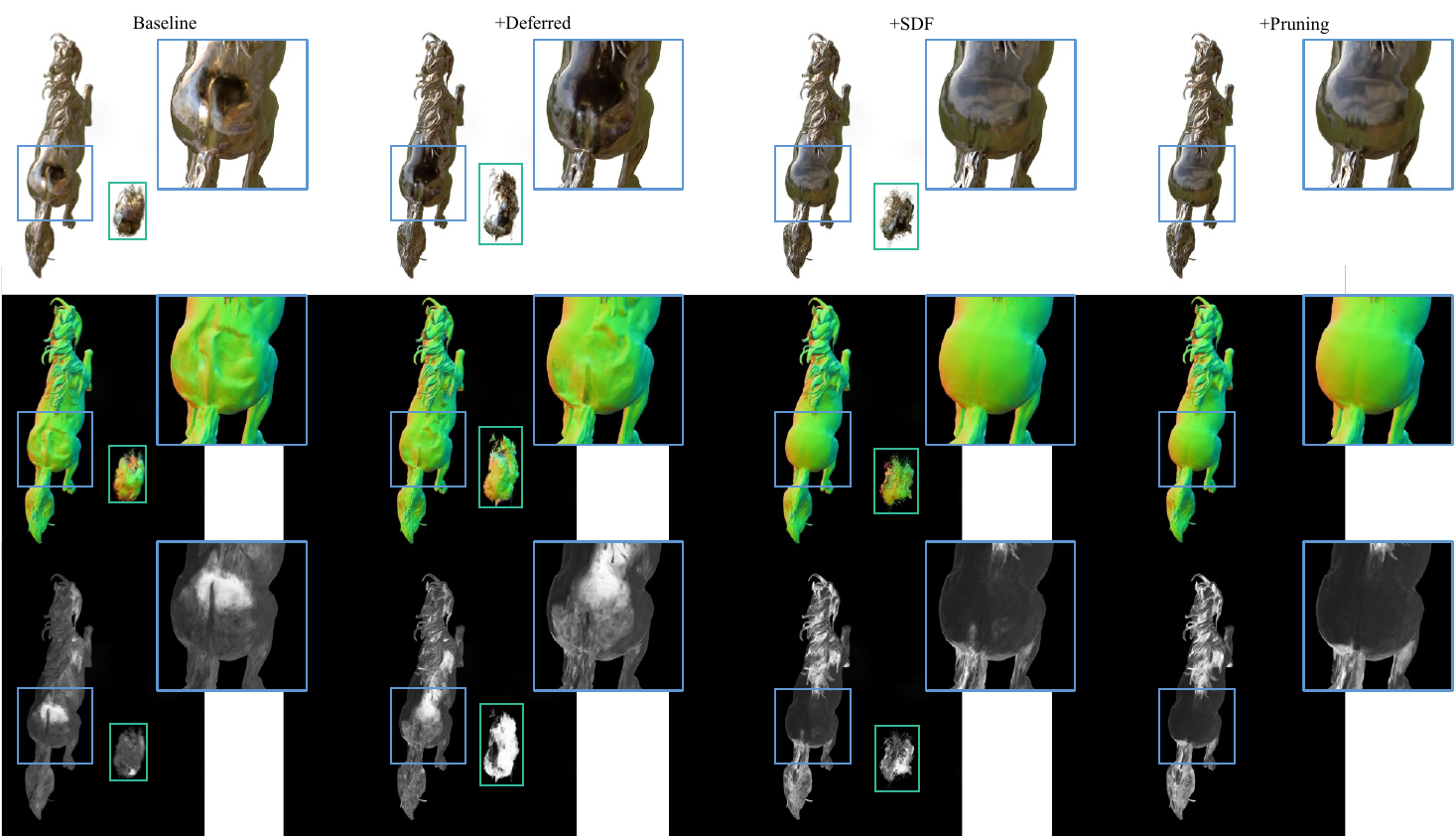}
    \caption{Ablation of several key components in our method, including deferred shading, SDF supervision, and pruning. The first, second, and third rows show relighting results, normal, and roughness, respectively. The insets are shown in blue, and some floaters are shown in green.}
    \Description[Ablation of several key components in our method]{The relighting results, normal and roughness visualizations of four variants. Starting from baseline with forward rendering, we apply deferred shading, the SDF priors, and the pruning step by step. Baseline cannot render specular highlights, provide problematic normals, and roughness maps. Deferred shading promotes the quality of highlights. SDF priors lead to reasonable decomposition. The pruning diminishes the floaters.}
    \label{fig:ablation}
\end{figure*}

\paragraph{NVS}
We also present the NVS results on the Glossy Blender dataset and the Shiny Blender Dataset. The quantitative evaluation in terms of PSNR and SSIM is presented in Tables \ref{tab:nvs_glossy} and \ref{tab:shiny_nvs_psnr}. Our method can provide high-quality NVS results with sharper highlights, which are present in Figures \ref{fig:nvs_glossy} and \ref{fig:nvs_shiny}, revealing the wide applications of our method. Our method outperforms the existing Gaussian-based methods on most scenes. However, our method fails in the Coffee scene, as the deferred pipeline assumes an opaque surface, and the liquid in the Coffee does not meet the assumption.

\paragraph{Geometry quality}

We present the comparisons of geometry quality on the Shiny Blender and Glossy Blender datasets. The choice of measurement is based on the available ground truth. The normal estimation on the Shiny Blender dataset in \figref{fig:normal_cmp_shiny} and MAE in \tabref{tab:shiny_mae}. Thanks to the SDF priors, our method can provide robust normal estimation, while the results of other Gaussian-based methods are overfit or noisy.

The CD on the Glossy Blender dataset is in \tabref{tab:nero_cd_gs}. Our method achieves the best reconstruction quality regarding CD among the Gaussian-based methods. These Gaussian-based methods have no specific design for reflective objects, and some methods (e.g., PGSR) assume multi-view consistency of appearance, which does not hold for reflective objects. Hence, all these methods perform poorly on the Glossy dataset. Although GSDF is most relevant to our method, it does not converge for reflective objects even with mask loss, and thus its result is not available. 
Besides, we show the mesh visualizations of representative methods in \figref{fig:mesh_glossy}. Visually, the existing Gaussian-based methods cannot provide smooth surfaces. TensoSDF and NeRO can provide smooth meshes while discarding some geometry details (see the mane of `horse'). Our method can provide globally smooth meshes with geometry details.



\subsection{Ablation study}
\label{sec:ab}


We conduct ablation studies on the key components of our model on the Glossy Blender dataset. The mean relighting PSNR and SSIM are presented in the \tabref{tab:ab}. We can see a consistent improvement when we employ new components to the model. We start from a baseline that first computes colors for Gaussians and alpha-blends them as in 3DGS. As shown in the 1st column in \figref{fig:ablation}, the baseline cannot render a reflective surface and provide the result with obvious artifacts due to the blending issue discussed in \secref{sec:defer}.

\paragraph{Deferred Gaussian splatting pipeline}
After applying the deferred Gaussian splatting pipeline, this variant provides a more specular result. However, due to the erroneous normal, the surface reflects light from the wrong directions, shown in the 2nd column of \figref{fig:ablation}. We observe more artifacts from the deferred results, for the pixel color from deferred shading corresponds to one shading point on the estimated surface while the one from forward shading is a combination of rendered colors of multiple Gaussians. The deferred formulation leaves less freedom for individual Gaussians to overfit. Consequently, more Gaussian outliers are generated to overfit sharp highlights, causing additional artifacts. Nevertheless, the deferred shading is necessary for specular materials (\secref{sec:defer}), and its issue has been solved in our full model.

\begin{figure}[tb]
    \centering
    \includegraphics[width = \linewidth]{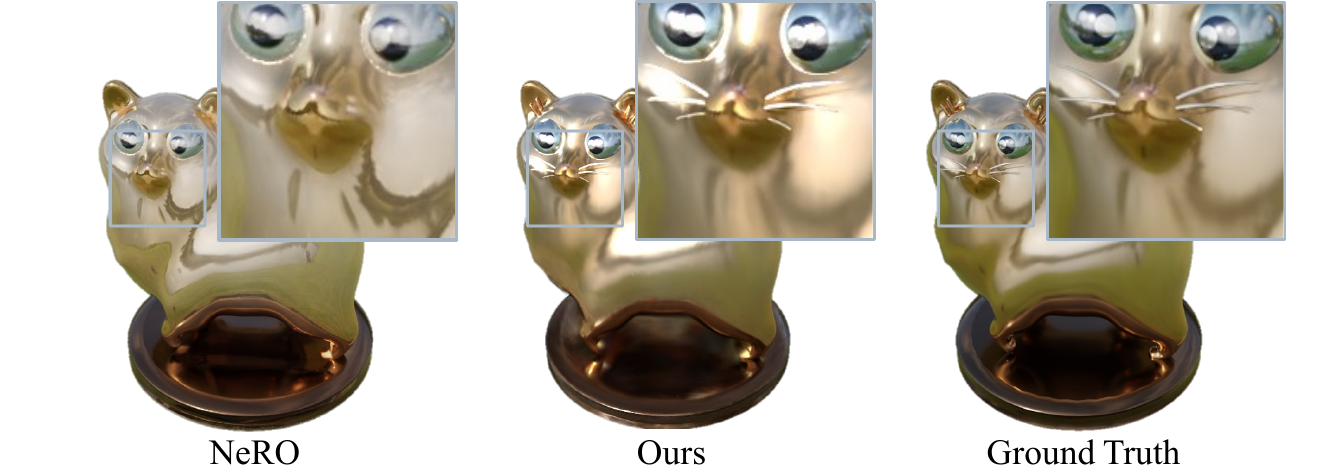}
    \caption{Our method preserves the details that are lost in the SDF-based method (i.e., NeRO), revealing the benefits of the Gaussian representation.}
    \Description[More details preserved by our method]{The comparison between NeRO and GS-ROR. Our method reconstructs the cat whiskers while NeRO fails.}
    \label{fig:sdfvsgs}
\end{figure}

\paragraph{TensoSDF incorporation}
Then, we incorporate the TensoSDF and utilize its priors to regularize the geometry. The model without TensoSDF is vulnerable to local minima and overfits under the training light condition, leading to erroneous geometry estimation. After employing the TensoSDF (3rd column in \figref{fig:ablation}), the model mitigates the ambiguity between material and geometry, thus predicting reasonable normal and material parameters (\eg roughness). We validate the effectiveness of our mutual supervision by comparing it to supervision by a pretrained TensoSDF. As shown in \tabref{tab:sdf_pre}, although using the pretrained TensoSDF for supervision could improve the quality slightly, it needs 5 hours for TensoSDF training. Optimizing GS with full-resolution SDF, as in GSDF, also improves the performance, increasing 40\% (+8.5GB) memory usage. In contrast, our method achieves a balance between performance and efficiency. Besides, our method preserves the details while SDF-based methods fail, as shown in \figref{fig:sdfvsgs} (see the cat whisker), revealing the benefits of our Gaussian representation.

\begin{table}[tb]
    \centering
    \caption{Ablation study of three key components on the Glossy Blender dataset. ``Def.'' means the deferred Gaussian splatting pipeline, ``Inc.'' means the incorporation of SDF and Gaussian, and ``Pru.'' means the pruning operation.}
    \resizebox{0.99\linewidth}{!}{
    \renewcommand{\arraystretch}{1.2}
    \begin{tabular}{ccc|ccc}
        \hline
        \multicolumn{3}{c|}{Components} & \multicolumn{3}{c}{Scene} \\
       \hline
         \multirow{2}{*}{Def.} & \multirow{2}{*}{Inc.} & \multirow{2}{*}{Pru.}               & Angel       & Cat         & Horse  \\
                &       &                  & PSNR/SSIM & PSNR/SSIM & PSNR/SSIM  \\
          
       \hline
                  &           &            & 18.86/0.8562 & 24.18/0.9253 & 20.33/0.9012 \\
        \ding{52} &           &            & 18.86/0.8574 & 24.87/0.9323 & 20.59/0.9089 \\
        \ding{52} & \ding{52} &            & \subsota{20.48/0.8793} & \subsota{26.22/0.9417} & \subsota{23.12/0.9357} \\
        \ding{52} & \ding{52} & \ding{52}  & \sota{20.81/0.8775} & \sota{26.28/0.9421} & \sota{23.31/0.9376} \\ 
       \hline
    \end{tabular}
    }
    \label{tab:ab}
\end{table}

\begin{table}[tb]
    \centering
    \caption{Comparison of the relighting quality between mutual supervision (Ours), using a pretrained TensoSDF to supervise Gaussians unidirectionally (Pretrained), and jointly optimization of full-resolution SDF with GS (Joint opt.) as in GSDF on the Glossy Blender dataset. Despite the slight improvement, using the pretrained model and the joint optimization leads to an extensive training time and memory cost, respectively.}
    \vspace{-0.2cm}
    \resizebox{0.99\linewidth}{!}{
    \renewcommand{\arraystretch}{1.2}
    \begin{tabular}{c|ccc}
        \hline
                          & Angel       & Cat         & Horse  \\
       \hline
        Ours       & 20.81 & 26.28 & 23.31  \\
        Pretrained & 21.45 (+0.64/+5h) & 26.37 (+0.09/+5h) & 24.59 (+1.28/+5h) \\
        Joint opt. & 21.68 (+0.87/+8.5GB) & 26.50 (+0.22/+8.5GB) & 24.34 (+1.03/+8.5GB) \\
       \hline
    \end{tabular}
    }
    \label{tab:sdf_pre}
\end{table}

\begin{figure}[tb]
    \centering
    \includegraphics[width = \linewidth]{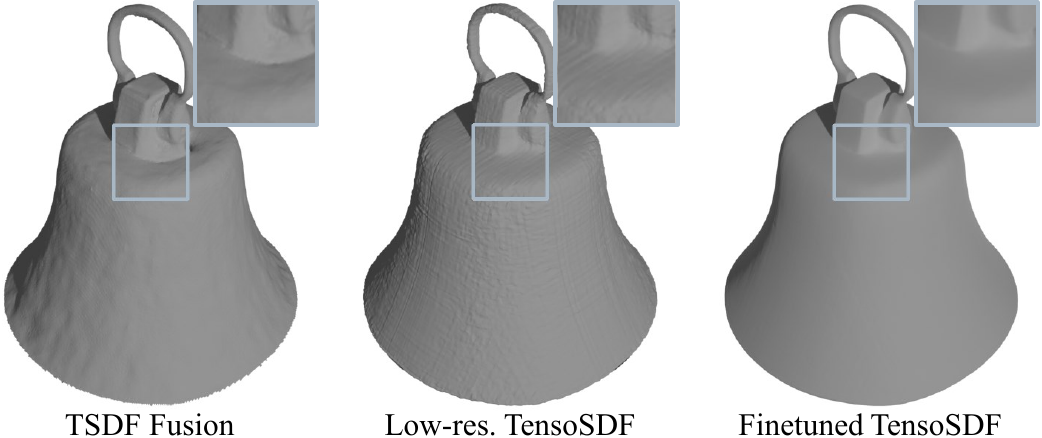}
    \caption{The reconstruction results from TSDF fusion, low-res. TensoSDF and finetuned TensoSDF. The meshes from TSDF fusion and low-res. TensoSDF has obvious artifacts, while our finetuned TensoSDF provides globally smooth meshes with fine-grained details.}
    \Description[Comparison of meshes extracted in different ways.]{The extracted meshes from TSDF fusion and low-resolution TensoSDF show obverse bumpy artifacts, while the finetuned TensoSDF produces smooth meshes.}
    \label{fig:ab_ft}
\end{figure}

\paragraph{SDF-aware pruning}
Eventually, we apply the SDF-aware pruning operation to our model. The models without pruning are likely to overfit under training views, and we can observe some unnecessary Gaussians under test views, causing severe floaters. Note that the overfitted Gaussians are invisible and in the mask region in training views. Therefore, we still observe the floater in the testing views, even though we apply the mask loss during training. After applying pruning, all Gaussians are located near the surface at the end of the training, and the model provides relighting results without floater.

\paragraph{GS-guided TensoSDF enhancement}
The above designs ensure a promising normal for relighting, while the enhancement enables high-quality geometry reconstruction. We show the visualization of meshes from TSDF fusion, low-resolution TensoSDF, and finetuned TensoSDF in \figref{fig:ab_ft}. Due to the inconsistency between the normal and depth of Gaussian primitives, although the blended normal is smooth, artifacts are apparent in the mesh from TSDF fusion. Furthermore, the low-resolution TensoSDF cannot provide smooth meshes and leads to grid-like artifacts. After finetuning, the upsampled TensoSDF provides globally smooth meshes with fine-grained details. We measure the mesh quality regarding CD in \tabref{tab:mesh_ab}, and our finetuned TensoSDF provides the best performance. 

\begin{table}[tb]
    \centering
    \caption{Comparison of the mesh quality regarding CD$\downarrow$. (CD is multiplied by $10^2$) We compare three variants: the mesh of TSDF fusion from Gaussians, the mesh from low-resolution TensoSDF, and our finetuned TensoSDF.}
    \vspace{-0.2cm}
    \begin{tabular}{c|ccc}
        \hline
                          & Angel       & Cat         & Horse  \\
       \hline
        TSDF Fusion       & 0.63 & 2.22 & 0.61 \\
        Low-res. TensoSDF  & \subsota{0.61} & \subsota{1.35} & \subsota{0.44} \\
        finetuned TensoSDF & \sota{0.41} & \sota{1.34} & \sota{0.34} \\
       \hline
    \end{tabular}
    \label{tab:mesh_ab}
\end{table}

\subsection{Discussion and limitations}
\label{sec:lim}

\begin{figure}[tb]
    \centering
    \includegraphics[width = \linewidth]{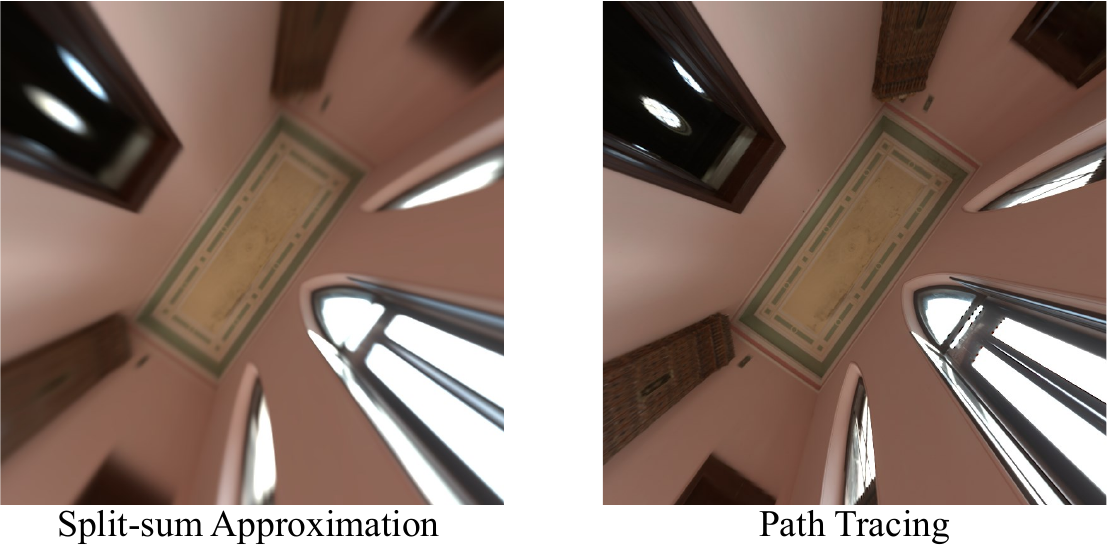}
    \caption{The comparison between the result using the Cycle path-tracing renderer in Blender and the one using split-sum approximation in DiffRast. The split-sum result is blurry with extremely low roughness. }
    \Description[Limitation of the split-sum approximation.]{Compared to the path tracing, the result of a low-roughness surface from the split-sum approximation is blurry.}
    \label{fig:limit_ss}
\end{figure}

\begin{figure}[tb]
    \centering
    \includegraphics[width = \linewidth]{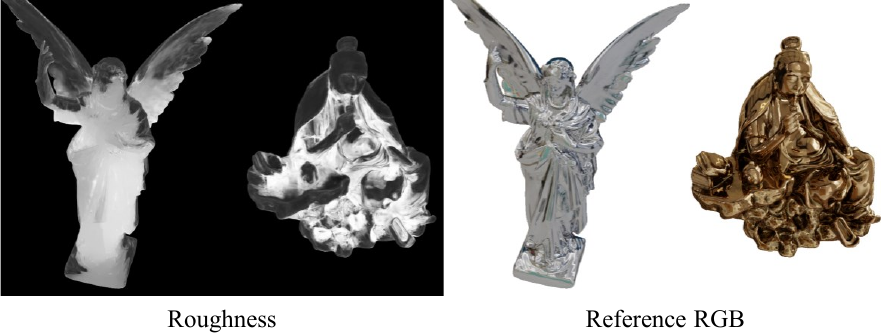}
    \caption{Due to the lack of indirect illumination modeling, our method fails to decompose reasonable material parameters (\eg roughness) on the concave surface with complex inter-reflection.}
    \Description[Limitation from the lack of illumination modeling]{Due to the lack of indirect illumination modeling, the decomposed roughness in the concave regions is problematic.}
    \label{fig:limit_ind}
\end{figure}

In this part, we discuss several key differences between our method and other SDF-based methods. 
We mainly discuss two groups of closely related methods. The first group is the ones that optimize GS and SDF jointly (\eg NeuSG, GSDF), and the second is the volumetric SDF methods. Then, we demonstrate some limitations of our methods and their potential solutions.

\paragraph{GS with SDF} Existing methods that optimize GS with SDF are designed for surface reconstruction and model the appearance by radiance, thus not supporting relighting. Due to the simplified appearance model, these methods fail to reconstruct reflective objects. 
NeuSG links SDF with GS only by aligning their normal. SDF and GS benefit marginally from each other, so it still takes 16 hours on an RTX 4090 for training. GSDF accelerates ray sampling of SDF by Gaussian depth and controls Gaussian opacity by SDF value. \revised{However, their method optimizes a full-resolution SDF with GS, which still takes 2 hours on an RTX A100 for training. On the contrary, we utilize a low-resolution TensoSDF to regularize GS with mutual supervision, which is more efficient.} Additionally, we use the normal from fixed GS to enhance the geometry details of full-resolution TensoSDF efficiently. Note that though DeferredGS introduces SDF into GS, it only uses the pretrained SDF to regularize GS, while SDF cannot benefit from GS. 

\paragraph{Volumetric SDF} Another group is the volumetric SDF-based methods that support relighting. As discussed in related works, these methods take a long time to converge, especially for the ones that handle reflective objects. We compare the open-source methods in the experiment section, while Neural-PBIR does not release the code. Its idea of introducing differentiable path tracing to model complex inter-reflection could benefit the Gaussian-based methods.

\paragraph{Limitations} Our model improves the relighting and mesh quality compared to other Gaussian-based methods. However, some issues remain to be solved. 
We observe that the split-sum approximation in DiffRast~\cite{laine_2020_diffrast} causes blurry renderings for low-roughness surfaces.  We render a mirror-like plane using the Cycle path-tracing renderer in Blender and the split-sum approximation in the DiffRast. The comparison is in \figref{fig:limit_ss}, and the split-sum result is blurry. We believe replacing it with more accurate rendering methods will improve the relighting quality. 
Last, we only consider direct lighting in our framework, which causes inconsistent material decomposition in the regions with complex inter-reflection, as shown in \figref{fig:limit_ind}. Introducing indirect illumination will benefit the relighting quality. We leave it for future work.

\section{Conclusion}

In this paper, we present a novel framework for real-time reflective object inverse rendering. We design an SDF-aided Gaussian splatting framework, using the mutual supervision of the depth and normal between deferred Gaussians and SDF to improve the geometry quality from Gaussians. Besides, we propose an SDF-aware pruning strategy with an automatically adjusted threshold, regularizing the position of Gaussian and avoiding the floater artifact. Finally, we design a GS-guided SDF enhancement to extract high-quality mesh by finetuning the TensoSDF.
Consequently, our method outperforms the existing Gaussian-based inverse rendering methods without losing efficiency. It is competitive with the NeRF-based methods in terms of relighting and mesh quality, with much less training and rendering time. In future work, extending our framework to complex scenarios with multiple objects is a potential direction.

\begin{acks}
The authors would like to thank the editor and the reviewers for their critical and constructive comments and suggestions. This work has been partially supported by the National Science and Technology Major Project under grant No. 2022ZD0116305, National Natural Science Foundation of China (NSFC) under grant No. 62172220, and the NSFC under Grant Nos. 62361166670 and U24A20330. Computation is supported by the Supercomputing Center of Nankai University.
\end{acks}


\bibliographystyle{ACM-Reference-Format}
\bibliography{paper}
\includepdf[pages=-]{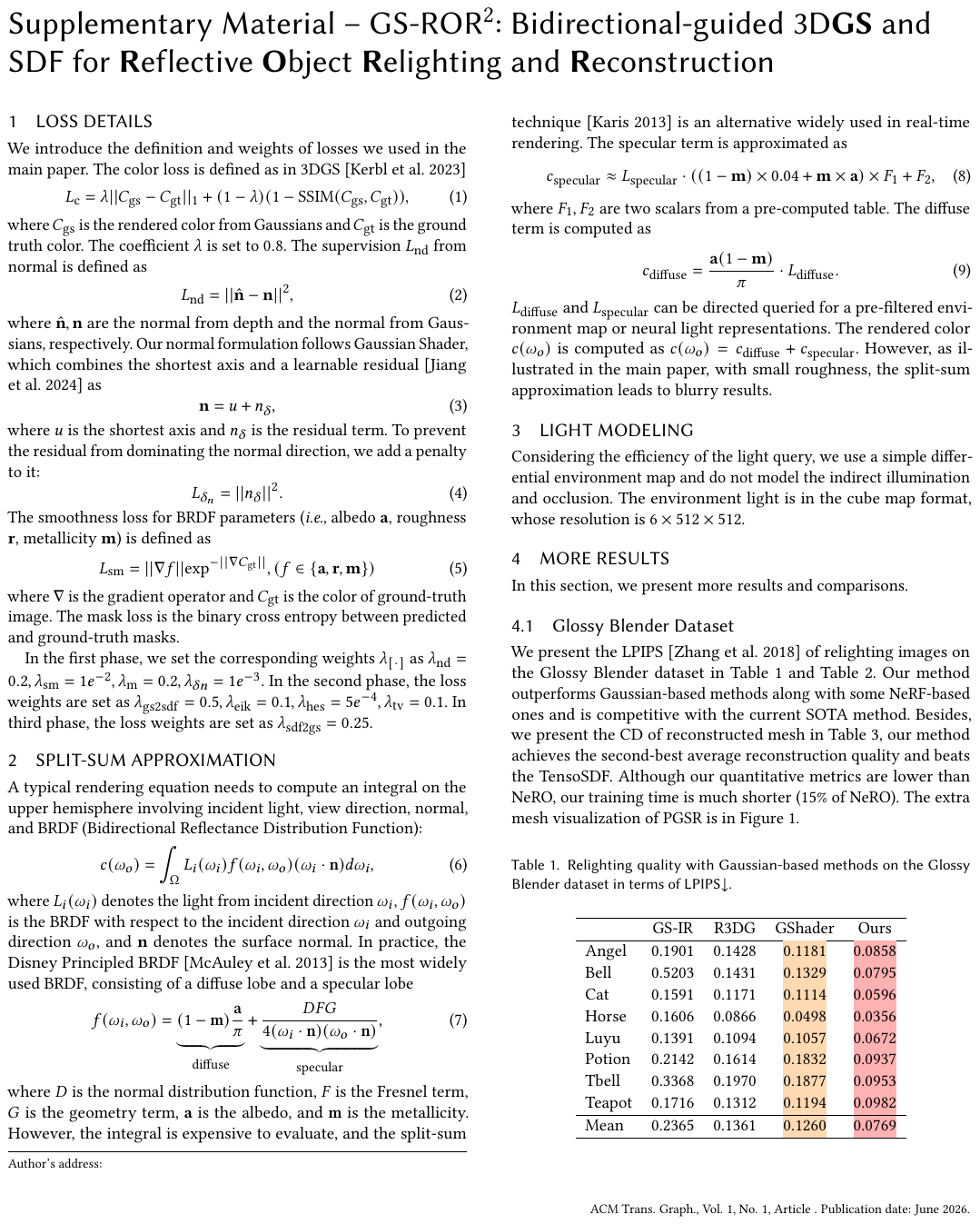}

\end{document}